\documentclass[a4paper,fleqn]{cas-dc}

\usepackage[numbers]{natbib}

\usepackage{amsmath,amsfonts}
\usepackage[ruled,vlined]{algorithm2e}
\usepackage{array}
\usepackage{textcomp}
\usepackage{stfloats}
\usepackage{url}
\usepackage{verbatim}
\usepackage{graphicx}
\usepackage{xcolor}

\usepackage{multirow}
\usepackage{multicol}
\usepackage{mathtools,xparse}
\usepackage{amssymb}
\usepackage{array}
\usepackage{colortbl}
\usepackage{bbding}
\usepackage{booktabs}

\def\tsc#1{\csdef{#1}{\textsc{\lowercase{#1}}\xspace}}
\tsc{WGM}
\tsc{QE}

\newcommand{\hv}[0]{\ensuremath{\boldsymbol{h}} }

\newcommand{\tv}[0]{\ensuremath{\boldsymbol{t}} }

\newcommand{\xv}[0]{\ensuremath{\boldsymbol{x}} }

\newcommand{\zv}[0]{\ensuremath{\boldsymbol{z}} }

\newcommand{\NonHangingInput}[2]{%
  \par\noindent\textbf{#1:} #2\par%
}

\begin{document}

\title [mode = title]{Semantically Guided Dynamic Visual Prototype Refinement for Compositional Zero-Shot Learning}                      



\author[1,2]{Zhong Peng}[style=chinese]\cormark[1]
\ead{mnspz@stu.xidian.edu.cn}
\author[1]{Yishi Xu}[style=chinese]\cormark[1]
\ead{xuyishi@stu.xidian.edu.cn}
\author[1,2]{Gerong Wang}[style=chinese]
\author[1]{Wenchao Chen}[style=chinese]
\author[1]{Bo Chen}[style=chinese]\cormark[2]
\ead{bchen@mail.xidian.edu.cn}

\author[2]{Jing Zhang}[style=chinese]\cormark[2]
\ead{ben_bbzj@126.com}
\author[1]{Hongwei Liu}[style=chinese]

\affiliation[1]{organization={National Key Laboratory of Radar Signal Processing, Xidian University},
city={Xi’an},
country={China}}

\affiliation[2]{organization={Research Institute of Systems Engineering, 	
Academy of Military Science},
city={Beijing},
country={China}}

\cortext[cor1]{Contributed equally}
\cortext[cor2]{Corresponding authors}

\begin{abstract}
Compositional Zero-Shot Learning (CZSL) seeks to recognize unseen state–object pairs by recombining primitives learned from seen compositions. Despite recent progress with vision–language models (VLMs), two limitations remain: (i) text-driven semantic prototypes are weakly discriminative in the visual feature space; and (ii) unseen pairs are optimized passively, thereby inducing seen bias. To address these limitations, we present \emph{Duplex}, a framework that couples dual-prototype learning with dynamic local-graph refinement of visual prototypes. For each composition, \emph{Duplex} maintains a semantic prototype via prompt learning and a visual prototype for unseen pairs constructed by recombining disentangled state and object primitives from seen images. The visual prototypes are updated dynamically through lightweight aggregation on mini-batch local graphs, which incorporates unseen compositions during training without labels. This design introduces fine-grained visual evidence while preserving semantic structure. It enriches class prototypes, better disambiguates semantically similar yet visually distinct pairs, and mitigates seen bias. Experiments on MIT-States, UT-Zappos, and CGQA in closed-world and open-world settings achieve competitive performance and consistent compositional generalization. Our source code is available at https://github.com/ISPZ/Duplex-CZSL
\end{abstract}


\maketitle

\section{Introduction}
Compositional generalization is an important yet complex cognitive ability~\cite{Graph-TPAMI, 2025zhangPAMI}. It allows humans to recombine familiar concepts to interpret or produce novel ones~\cite{lake2023human, ijcai2024_Zhang}. For example, upon encountering ``\textit{black swan}'' for the first time, one can recognize it by integrating prior knowledge of the category ``\textit{swan}'' with the state ``\textit{black}.'' Compositional Zero-Shot Learning (CZSL)~\cite{2017CVPRLE+, 2023PAMI-Simple-Primitives, Kim_2023_ICCV, li2020symmetry, CANET_2023CVPR} aims to develop models that recognize unseen state–object compositions although the training data contain only seen state–object pairs. {Consistent with common practice, we use the term ``state'' as a unified abstraction for attribute-like visual modifiers (e.g., texture, material, or color) in standard benchmarks.} The central objective is to decouple state and object representations from observed compositions and then recombine them to identify previously unobserved pairs, thereby alleviating the need to enumerate all state–object compositions. 

Existing CZSL approaches~\cite{2017CVPRLE+, nagarajan2018attributes, Purushwalkam_2019_ICCV, PROCC_AAAI2024, li2020symmetry, mancini2021open, naeem2021learning, Graph-TPAMI} primarily follow two paradigms: (i) \emph{joint representation learning}, which embeds visual state–object compositions and textual semantics into a shared space and constrains this space via alignment strategies such as graph-embedding methods~\cite{naeem2021learning, Graph-TPAMI, kipf2016semi, park2020sumgraph} and semantic transformation~\cite{nagarajan2018attributes} to transfer composition rules to unseen pairs; and (ii) \emph{disentanglement-based methods}, which employ separate modules (e.g., cross-attention~\cite{hao2023ade} or a multi-layer perceptron (MLP)~\cite{invariant_2022ECCV}) to model state and object features independently~\cite{Li_2022_CVPR, Saini_2022_CVPR}. Both paradigms often rely on features from pretrained visual encoders and static word embeddings such as GloVe or word2vec, which can limit effective modeling and transfer to unseen compositions. These limitations have motivated the community to leverage vision–language models (VLMs) that provide stronger cross-modal alignment.

With the strong cross-modal alignment of VLMs such as CLIP~\cite{pmlr-v139-radford21a}, recent work has adopted prompt-tuning strategies~\cite{csp2023, GIPCOL, Lu_2023_CVPR, Huang2024Troika, xu2022prompting, 2024CVPR-CDSCZSL} to enhance compositional reasoning, including compositional soft prompts~\cite{csp2023}, decomposed textual features~\cite{Lu_2023_CVPR}, and cross-modal fusion~\cite{Huang2024Troika}. {Representative methods such as CSP~\cite{csp2023}, DFSP~\cite{Lu_2023_CVPR}, GIPCOL~\cite{GIPCOL} follow this semantic-centric design. They primarily adapt CLIP by learning prompts or decomposed textual embeddings while keeping the visual encoder largely fixed, so the prototypes for classification reside mainly in the semantic space. As a result, visual evidence is incorporated only implicitly via alignment to text-driven prototypes, without explicitly modeling or updating visual composition prototypes.} Although VLMs provide a reliable joint vision-language space for concept generalization, two key challenges remain. \textbf{(1) Limited discriminability and adaptability of semantic prototypes.} Prototype-based VLM methods often learn semantic prototypes to as surrogates for unseen visual concepts; however, textual prototypes may lack the fine-grained visual details needed to distinguish closely related compositions, blurring decision boundaries in the visual space. \textbf{(2) Passive treatment of unseen compositions and ``seen bias.''} Because unseen compositions lack annotations, they are commonly excluded from training dynamics, leaving only passive transfer at inference. Training gradients are typically dominated by seen compositions, which can shift prototypes and decision boundaries toward seen classes, widen the training and test distribution gap, and reduce the separability of unseen compositions. {These limitations are particularly pronounced in the semantic-centric pipelines above. The absence of explicitly maintained visual composition prototypes and the lack of training-time interaction with unlabeled unseen compositions make it difficult to recalibrate decision boundaries once strong seen co-occurrence patterns have been learned.} In this work, we move beyond purely semantic optimization. We instead preserve semantic prototypes as stable, interpretable anchors and refine only the visual prototypes that determine fine-grained decision boundaries, {forming a dual set of prototypes (semantic and visual) and updating the visual prototypes through an active, label-conditioned refinement process that is explicitly guided by the semantic anchors.}

\begin{figure}[pos=!t]
    \centering
    \includegraphics[width=1.0\linewidth]{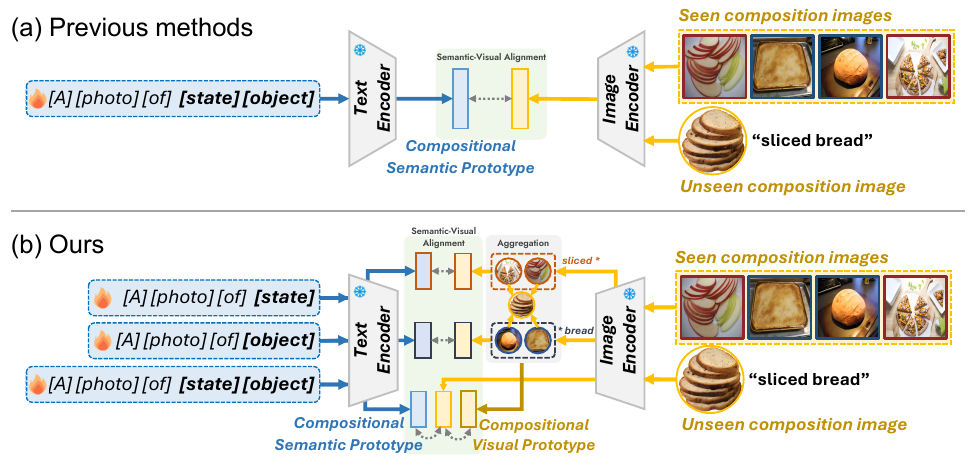}
    \caption{(a) Previous methods learn a \emph{single} compositional \emph{semantic} prototype from textual prompts and align images only via semantic supervision—interpretable but short on fine-grained visual cues, treating unseen compositions passively. 
    (b) \emph{Duplex} maintains \emph{semantic and visual} compositional prototype for each state–object pair: visual prototypes are formed by decoupling seen states/objects and counterfactual recomposition, then \emph{actively} refined on mini-batch via lightweight aggregation. This injects visual evidence and mitigates seen-bias, improving generalization to unseen compositions.}
    \label{fig:motivation}
    \vspace{-5mm}
\end{figure}

To address the above issues, we present \emph{Duplex}, a CZSL framework that couples dual-prototype learning with \emph{active local-graph updates of the visual prototypes} (Fig.~\ref{fig:motivation}). The framework maintains both a semantic prototype and a visual prototype for each composition and iteratively refines the visual prototype through dynamic aggregation on mini-batch local graphs, thereby incorporating unseen compositions into training. \emph{Semantic prototype.} Prompt learning forms learnable compositions of state and object textual embeddings, preserving interpretable semantic structure and providing anchors for cross-modal alignment. These anchors impose semantic-consistency constraints that condition local-graph aggregation and steer visual prototype updates toward compositional semantics. 
\emph{Visual prototype.} From seen samples, states and objects are explicitly disentangled in the visual feature space and recombined counterfactually to approximate potential unseen compositions. A local graph is then constructed within each mini-batch, and lightweight graph aggregation, such as GCN message passing with consistency constraints, is applied for progressive refinement. All refinements occur during training. At inference, \emph{Duplex} retrieves from the global codebook without additional computation. Compared with global aggregation, local aggregation suppresses spurious cross-class shortcuts, reduces intra-class variance, and preserves inter-class boundaries. 

Overall, \emph{Duplex} retains the interpretability of semantic anchors while injecting fine-grained visual evidence and continuously calibrating representations for unseen compositions.  Semantically guided refinement of visual prototypes tightens clusters without collapsing decision boundaries, attenuates seen-dominant bias, and aligns visual updates with the textual compositional structure, thereby narrowing the training and test gap.

\textbf{In a nutshell, our contributions are threefold.}
\begin{itemize}
\item We identify two persistent bottlenecks in CZSL, even with VLM support. The first is \emph{semantic projection bias}, which constrains fine-grained separability. The second is \emph{seen-dominant optimization}, which induces seen bias and limits the engagement of unseen pairs in training dynamics.
\item We introduce \emph{Duplex}, which learns semantic prototypes via prompt learning and constructs visual prototypes by disentangling state and object features from seen samples and recombining them counterfactually. The visual prototypes are then actively updated through dynamic aggregation on mini-batch local graphs, enabling training without unseen labels.
\item Experiments on MIT-States, UT-Zappos, and CGQA under closed-world (CW) and open-world (OW) settings demonstrate competitive performance and compositional generalization across datasets.
\end{itemize}

\section{Related Work}

\subsection{Compositional Zero-Shot Learning}

{Zero-Shot Learning (ZSL) has evolved into a diverse field that encompasses paradigms such as generative modeling, multi-modal transfer, and different embedding strategies, as detailed in surveys on multimodal zero-shot learning and zero-shot object detection~\cite{cao2023review,cao2025survey}. 
Within this broader landscape, \textbf{Compositional Zero-Shot Learning (CZSL)} is recognized as a distinct and challenging subfield~\cite{Graph-TPAMI,ZHANG2025130213,HAO2025128773}. Unlike standard ZSL, CZSL places stronger emphasis on the \emph{compositional mechanisms} that enable generalization to novel state-object configurations. It requires models to disentangle and recombine primitive concepts rather than merely mapping holistic class labels.}

Early efforts in CZSL primarily followed one of two foundational strategies. The first projects the textual labels of state-object pairs and their corresponding visual features into a shared embedding space~\cite{2017CVPRLE+}. Within this paradigm, a range of techniques have been explored, including modeling states as linear transformations applied to object embeddings~\cite{nagarajan2018attributes}, hierarchically decomposing and recomposing visual primitives~\cite{yang2020learning, Khan_2023_WACV}, and assuming object representations to be invariant under state transformations~\cite{li2020symmetry, 2022MMCVGAE}. Others adopt compatibility-based formulations that model interactions among images, states, and objects~\cite{NEURIPS2020_1010cedf, Purushwalkam_2019_ICCV,2024TLVAR-CZSL}, with classification performed using similarity metrics in the joint embedding space. The second strategy uses dual-module architectures in which state and object classifiers are trained independently and their outputs are subsequently fused to yield compositional predictions. To improve generalization, several methods introduce disentanglement mechanisms~\cite{Li_2023_ICCV, ruis2021independent, hao2023ade, Saini_2022_CVPR}, while others incorporate external knowledge sources to suppress implausible or semantically inconsistent compositions~\cite{mancini2021open, Graph-TPAMI}.

Together, these approaches laid the conceptual and technical foundation for CZSL, while also revealing its central challenge, namely the combinatorial explosion of possible state-object pairs. Models trained solely on a limited subset of seen compositions tend to overfit to co-occurrence statistics and fail to generalize to novel combinations. This limitation highlights the inadequacy of purely data-driven paradigms and has motivated a growing consensus that effective CZSL models must integrate stronger structural priors and more expressive reasoning mechanisms.
{Relevant insights can also be drawn from the broader ZSL literature. Classical methodologies such as MFF~\cite{cao2022mff}, which performs multi-modal feature fusion in a shared embedding space, and graph-embedding based multi-label ZSL~\cite{zhang2023graph}, which encodes label relationships to transfer knowledge, exemplify the benefits of explicit structural priors. These principles suggest that capturing relational structure is key to robust compositional generalization.}

Large-scale vision-language models (VLMs)~\cite{pmlr-v139-radford21a, pmlr-v139-jia21b} align visual and textual representations during pre-training and have significantly reshaped the landscape of artificial intelligence. These models demonstrate strong performance across a range of downstream tasks, including visual question answering~\cite{shen2021much, clip_vqa}, image captioning~\cite{zeng2023conzic, mokady2021clipcap}, and zero-shot or few-shot image recognition~\cite{liu24uai, li2024graphadapter}. Currently, prompting techniques~\cite{zhou2022coop, NEURIPS2020_1457c0d6} are a primary approach for adapting VLMs to downstream tasks. Unlike fine-tuning the entire model, which is computationally expensive and prone to catastrophic forgetting, prompting achieves adaptation by introducing new inputs or small learnable modules.  In multimodal settings, prompts can take the form of fixed text templates or learnable embeddings~\cite{yang2021empirical}.

Recent advances address the unique challenges of CZSL along two complementary axes. First, prompting has been employed to extract and leverage the compositional priors implicitly encoded in pretrained VLMs such as CLIP~\cite{pmlr-v139-radford21a}. Second, explicit structural reasoning has been introduced through graph-based modeling. Prompt-based methods often replace fixed class labels with learnable state and object embeddings~\cite{csp2023}. For example, DFSP~\cite{Lu_2023_CVPR} integrates a cross-modal fusion module built on soft prompts, while GIPCOL~\cite{GIPCOL} employs prefix vectors inspired by algebraic structures such as group theory. Troika~\cite{Huang2024Troika} proposes a multi-path design to jointly model states, objects, and compositions, and CDS-CZSL~\cite{2024CVPR-CDSCZSL} introduces a diversity-driven specificity learning scheme to enhance discriminability.

However, most VLM-based CZSL pipelines remain \emph{semantic-centric}: they primarily optimize semantic prototypes (text embeddings) and align images to these anchors. Two gaps persist: (i) semantic prototypes, while interpretable, can be weakly discriminative in the visual space; (ii) unseen compositions are often optimized only \emph{passively} through semantic supervision, reinforcing seen bias. Our work addresses these gaps by keeping semantic prototypes as stable anchors while \emph{actively} refining \emph{visual} composition prototypes through label-conditioned local graphs built per mini-batch. This approach engages unseen compositions during training without additional labels and avoids the cost of global graphs.

\subsection{Prototype Learning} 
Prototype learning is central to few-shot recognition and has influenced both zero-shot and few-shot settings~\cite{2023CC-ZSL, NEURIPS2022Hou, NEURIPS2021_Wang, Hou_2024_CVPR, 2024DPN, 2023_IJCAI_HPL}. The key idea is to represent each class (state, object, or composition) by a representative embedding, with queries classified by similarity to class prototypes~\cite{2017nipsPrototypical}. In CZSL, prior work learns state/object prototypes and then composes or fuses them for novel pairs~\cite{li2020symmetry, Hu_Wang_2023}. {More recent methods refine prototypes further using graph-based interactions or contrastive objectives. For example, ProtoProp~\cite{ruis2021independent} constructs a static compositional graph over independent state and object prototypes and performs a single round of propagation decoupled from the training dynamics; 
SCEN~\cite{Li_2022_CVPR} learns state and object prototypes independently via contrastive learning and synthetic data augmentation, without explicitly modeling composition-specific prototypes on a graph; and ProtoLP~\cite{Zhu_2023_CVPR} builds graphs at inference time over support and query samples for post-hoc label propagation in a transductive few-shot setting.} While such designs bring interpretability and sample efficiency, prototypes are often static (derived once from text or pooled features) or updated only implicitly via the alignment loss, limiting their ability to capture fine-grained, context-dependent visual cues that distinguish hard compositions.

Our \emph{Duplex} framework differs in two respects. First, it maintains \emph{dual} prototypes at the composition level: semantic prototypes (prompt-learned, stable anchors) and visual prototypes (image-grounded, discriminative carriers). Second, it performs \emph{stepwise, local} refinement of \emph{visual} prototypes by aggregating instance-level evidence on a label-conditioned graph within each mini-batch. This actively injects fine-grained visual information into the prototype space and brings unseen compositions into the optimization loop, while preserving the semantic structure that supports generalization. {In contrast to existing graph-based prototype refinement methods, the graphs in Duplex are constructed \emph{during training}, on each mini-batch, with nodes that include disentangled state/object features, counterfactual composition features, and evolving visual prototypes. These graphs are conditioned on feasible state-object labels, allowing unseen compositions to be actively refined in the visual prototype space, while semantic prototypes, learned via CLIP prompts, provide stable vision-language anchors.} {Unlike prior semantic-centric CZSL methods, which optimize text-derived prototypes as the primary visual classifiers, our semantic prototypes act as globally learned but structurally stable anchors. They are updated only through prompt learning, whereas visual discriminability is maintained by image-grounded prototypes that are refined via local graph aggregation.} 
Finally, inference remains efficient by retrieving from a global codebook without additional computation.

\section{Methodology}
\label{sec:method}
\textbf{Problem Definition}
In compositional zero-shot learning (CZSL), we consider two sets of primitive concepts (states and objects) denoted by $\mathcal{S}=\{s_1,\dots,s_M\}$ and $\mathcal{O}=\{o_1,\dots,o_N\}$, with $M=|\mathcal{S}|$ and $N=|\mathcal{O}|$. The compositional label space is the Cartesian product $\mathcal{C}=\mathcal{S}\times\mathcal{O}$, so $|\mathcal{C}|=MN$. Each sample is associated with a compositional label $c=(s_m,o_n)\in\mathcal{C}$. We partition $\mathcal{C}$ into disjoint seen and unseen sets, $\mathcal{C}_s$ and $\mathcal{C}_u$, such that $\mathcal{C}_s\cap\mathcal{C}_u=\emptyset$ and $\mathcal{C}_s\cup\mathcal{C}_u=\mathcal{C}$. Training uses only labels from $\mathcal{C}_s$, while compositions in $\mathcal{C}_u$ are evaluated at test time. 

At inference, the target label set $\mathcal{C}_{\text{tgt}}$ defines two settings:
(i) \emph{Closed-world}: $\mathcal{C}_{\text{tgt}}=\mathcal{C}_s\cup\mathcal{C}_u'$, where $\mathcal{C}_u'\subset\mathcal{C}_u$ is a fixed, predefined subset of unseen compositions;
(ii) \emph{Open-world}: $\mathcal{C}_{\text{tgt}}=\mathcal{C}_s\cup\mathcal{C}_u=\mathcal{C}$, i.e., all state-object compositions.

\begin{figure*}[!t]
    \centering
    \includegraphics[width=\textwidth]{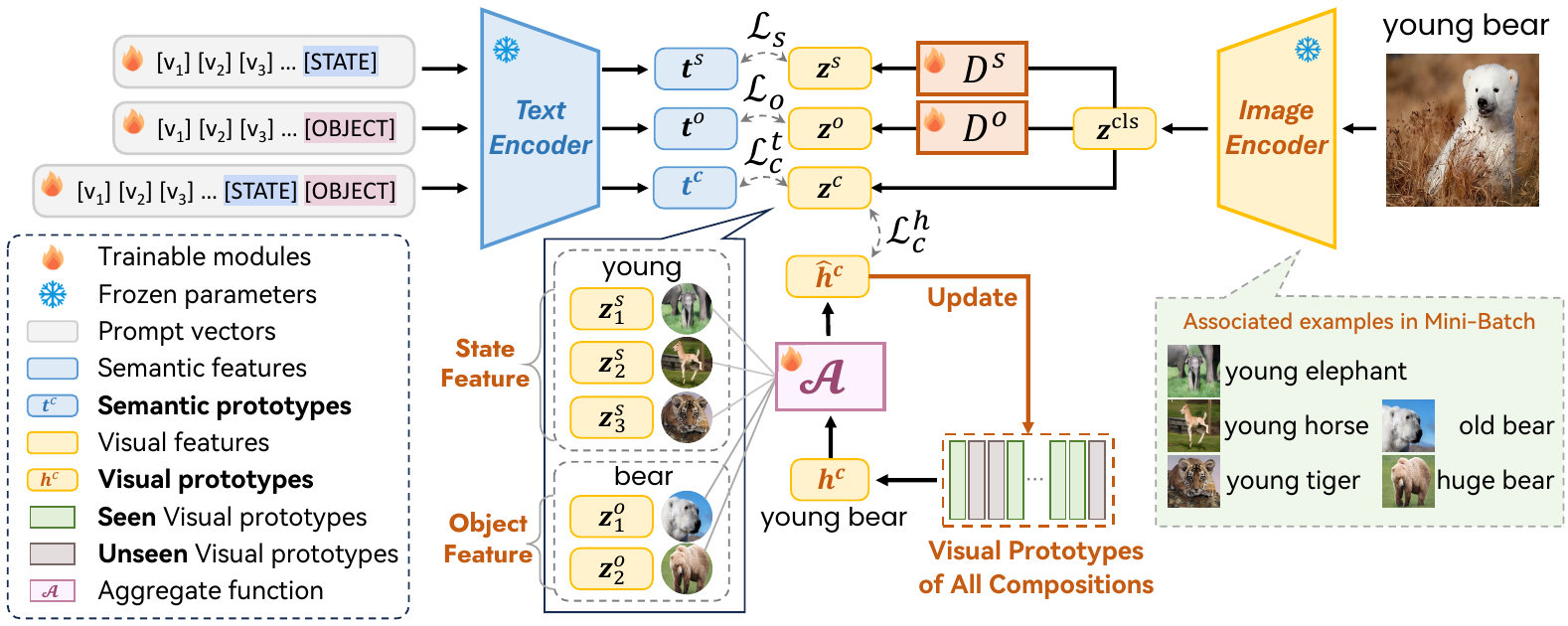}
    \caption{\textbf{Overview of \emph{Duplex}}. We learn (i) \emph{semantic prototypes} by composing state/object names with learnable soft prompts and a frozen CLIP text encoder, and (ii) \emph{visual prototypes} by disentangling image features into state/object factors, \emph{counterfactually recombining} them to cover unseen compositions, and \emph{refining} a global prototype codebook via a mini-batch \emph{local graph} with momentum updates. Dual prototypes serve as complementary anchors and are jointly aligned with image features.}
    \label{fig:CZSL}
\end{figure*}

\begin{figure}[pos=t]
\centering
\includegraphics[width=\linewidth]{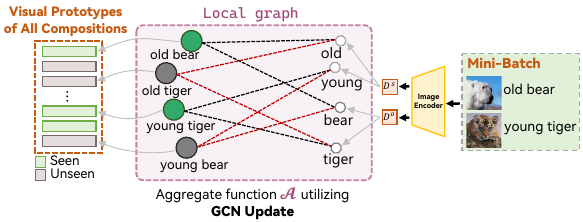}
\caption{\textbf{Local graph construction and visual prototype update.} Disentangled state/object factors in a mini-batch connect to semantically consistent prototype nodes (including unseen compositions). The aggregation function $\mathcal{A}(\cdot)$ in Eq.~\eqref{eq:aggregation} performs normalized, weighted message passing into prototype nodes, and a momentum update refines only the touched prototypes.}
\label{fig:local-graph}
\end{figure}

\textbf{Overall Framework} To tackle the challenge of Compositional Zero-Shot Learning (CZSL), we propose \emph{Duplex}, as illustrated in Fig.~\ref{fig:CZSL}. Our pipeline proceeds through four stages:
\textit{(1) Prompted text to semantic prototypes.}
We design three prompts for state, object, and state-object composition. Passing these prompts through a frozen text encoder yields a composition prototype $\tv^{c}$ and two textual factors $\tv^{s}$ and $\tv^{o}$. The prompt vectors are lightweight and trainable. Text-side objectives $\mathcal{L}_s, \mathcal{L}_o, \mathcal{L}_c^{t}$ maintain a well-structured semantic space.
\textit{(2) Image encoding and factorization.}
A frozen image encoder maps an input image to a global representation $\zv_{\text{cls}}$. Two shallow heads $D^{s}$ and $D^{o}$ project $\zv_{\text{cls}}$ into disentangled state and object factors $\zv^{s}$ and $\zv^{o}$. These factors are aligned with $\tv^{s}$ and $\tv^{o}$ using state and object alignment losses, encouraging modality consistency while keeping the backbone frozen.
\textit{(3) Global visual prototype codebook for all compositions.}
We maintain a visual prototype bank $\mathbf{H}=\{\hv^{c}\mid c=(s,o)\in\mathcal{C}_{\text{tgt}}\}$, allocating one prototype to every composition—including unseen pairs. Prototypes for seen pairs are initialized from data, while prototypes for unseen pairs are counterfactually composed from batchwise $(\zv^{s}, \zv^{o})$ factors. Classification over the global codebook with the prototype-classification loss $\mathcal{L}_{C}^{h}$ drives discriminative learning in the visual space.
\textit{(4) Dynamic visual-prototype refinement.}
For each target composition $c$ (e.g., ``young bear''), we build a label-conditioned local graph from associated examples in the current mini-batch (as shown in Fig.~\ref{fig:local-graph}, i.e., instances sharing the state or the object, such as ``young tiger'' or ``old bear''). A lightweight aggregation operator $\mathcal{A}$ pools their evidence to produce a refined visual prototype $\hat{\hv^{c}}=\mathcal{A}\!\big(\hv^{c};\text{neighbors}\big)$. The global codebook is then updated via a momentum write-back, so that refined prototypes continuously improve for both seen and unseen pairs. {Importantly, while semantic prototypes are globally optimized via prompt learning, they are explicitly exempted from local graph message passing. Within the local graph module, they serve as stable anchors that do not receive updates from visual neighbors. This design ensures that the learned semantic manifold, optimized over the entire dataset, is not distorted by batch-specific visual noise. Consequently, the local graph focuses exclusively on refining visual prototypes to align with these robust semantic references.}

At test time, \emph{Duplex} performs efficient nearest-prototype matching between $\zv_{\text{cls}}$ and the global codebook $\mathbf{H}$ without building any graph, preserving runtime efficiency while benefiting from the refined prototypes learned during training.

\subsection{Semantically Guided Prototype Construction}
\textbf{Prompted Semantic Prototypes.} 
Based on the success of previous works, learning effective semantic representations for compositions is essential to improving the performance of CZSL. The key challenge lies in constructing suitable prompts for compositional labels. Drawing on current best practices, we employ a fully learnable soft-prompt strategy. Specifically, we prepend a learnable soft prompt $\{\mathbf{v}_1^c,\ldots,\mathbf{v}_K^c\}$ to the tokenized state/object names to form
\begin{equation}
\mathbf{P}^c_{m,n}=\{\mathbf{v}_1^c,\ldots,\mathbf{v}_K^c,\,\boldsymbol{\rho}_m^{s},\,\boldsymbol{\rho}_n^{o}\},
\end{equation}
where \(\boldsymbol{\rho}_m^{s}\) and \(\boldsymbol{\rho}_n^{o}\) denote the token sequences of the \emph{state name} $s_m$ and the \emph{object name} $o_n$, respectively. Feeding $\mathbf{P}^c_{m,n}$ into the frozen CLIP text encoder $E_{txt}$ yields the \textbf{semantic prototype}
\begin{equation}\label{eq:semproto}
    \tv^c_{m,n}=E_{txt}(\mathbf{P}^c_{m,n}).
\end{equation}
With the semantic representations of compositions in hand, it is necessary to align them with the visual features of images so that the learned prompts generalize beyond the seen compositions. For images, we follow CLIP preprocessing and use ViT-L/14 as the image encoder $E_{img}$. The \texttt{[CLS]} token $\zv^\text{cls}$ serves as the global visual representation:
\begin{equation}\label{eq:imgfeat}
    \zv^\text{cls}_i=E_{img}(\xv_i),\qquad \zv^c_i\triangleq \zv^\text{cls}_i.
\end{equation}
According to Eq.~\ref{eq:semproto} and Eq.~\ref{eq:imgfeat}, we obtain the visual feature $\zv^c_i$ for the input image $\xv_i$ and the semantic representations of all compositions. This allows us to compute the probability that $\xv_i$ belongs to its corresponding compositional label $c_{m,n}^{t} = (s_m, o_n)$. The cross-entropy loss encourages the model to recognize the correct semantic role, and the formulation can be written as
\begin{equation}
\begin{aligned}    p(c_{m,n}^{t}\mid\xv_i)&=\frac{\exp(\zv^{c}_{i}\cdot\tv^c_{m,n}/\tau)}{\sum_{m=1}^{M}\sum_{n=1}^{N}\exp(\zv^{c}_{i}\cdot\tv^c_{m,n}/\tau)}, \\
\mathcal{L}^{t}_{c}&=-\frac{1}{|\mathcal{T}_{tr}|}\sum_{i}\log p(c_{m,n}^{t}\mid\xv_i).
\end{aligned}
\end{equation}
where $\tau\!\in\!\mathbb{R}$ is the predefined temperature parameter in CLIP. For simplicity, all features are $\ell_2$-normalized by default.

\textbf{State–Object Disentanglement of Visual Features.}
As discussed above, maintaining only semantic prototype representations for compositional labels can limit generalization. On the one hand, the model may overfit to seen compositional labels, as only their prompts are explicitly optimized. On the other hand, in terms of textual descriptions, the semantic differences among compositional labels may be subtle, making them difficult to separate in the representation space. Thus, a natural alternative is to seek visual prototype representations of compositions by leveraging the image information flow.

However, states and objects are intrinsically entangled in images. To learn rational visual prototypes for both seen and unseen compositions, we disentangle the state information from the object content. This design enables us to acquire visual prototype representations of any composition by later fusing the separate state and object features. Concretely, we introduce two disentanglers, $D^s$ and $D^o$, to separate the visual features of states and objects, respectively, from the global representation (\texttt{[CLS]} token) of the image:
\begin{equation}
\zv^s_i=D^s(\zv^\text{cls}_i),\qquad \zv^o_i=D^o(\zv^\text{cls}_i).
\end{equation}
where both $D^s(\cdot)$ and $D^o(\cdot)$ are implemented using two distinct two-layer MLPs.
It should be noted that merely inserting $D^s$ and $D^o$ does not automatically yield disentanglement. In practice, we need supervision that encourages specialization. Therefore, we construct prompts for states and for objects in a manner similar to compositions:
\begin{equation}
\mathbf{P}^s_m=\{\mathbf{v}_1^s,\ldots,\mathbf{v}_K^s,\,\boldsymbol{\rho}_m^{s}\},\quad
\mathbf{P}^o_n=\{\mathbf{v}_1^o,\ldots,\mathbf{v}_K^o,\,\boldsymbol{\rho}_n^{o}\}, 
\end{equation}
and obtain the corresponding $E_{txt}$ features
\begin{equation}
\tv^s_{m}=E_{txt}(\mathbf{P}^s_{m}),\qquad
\tv^o_{n}=E_{txt}(\mathbf{P}^o_{n}).
\end{equation}
With these semantic representations of states and objects, we align them with the corresponding visual features by jointly maximizing the probability of the true state or object to which the image belongs. This is formulated as
\begin{equation}
\begin{aligned}
p(s_m\mid\xv_i)&=\frac{\exp(\zv^{s}_{i}\cdot\tv^s_{m}/\tau)}{\sum_{m'=1}^{M}\exp(\zv^{s}_{i}\cdot\tv^s_{m'}/\tau)},\\
\mathcal{L}_s&=-\frac{1}{|\mathcal{T}_{tr}|}\sum_i\log p(s_m\mid\xv_i)
\end{aligned}
\end{equation}

\begin{equation}
\begin{aligned}
p(o_n\mid\xv_i)&=\frac{\exp(\zv^{o}_{i}\cdot\tv^o_{n}/\tau)}{\sum_{n'=1}^{N}\exp(\zv^{o}_{i}\cdot\tv^o_{n'}/\tau)},\\
\mathcal{L}_o&=-\frac{1}{|\mathcal{T}_{tr}|}\sum_i\log p(o_n\mid\xv_i).
\end{aligned}
\end{equation}
These auxiliary losses $\mathcal{L}_s$ and $\mathcal{L}_o$ drive $D^s$ and $D^o$ to retain information specific to states and objects, thereby providing controllable factors for compositions. Up to this point, we have found that our architecture aligns with the three-path paradigm proposed by~\cite{Huang2024Troika}. However, the two additional branches here are explicitly aimed at promoting efficient compositional learning of visual prototypes later on, as the instance-level state and object visual features are used to update the visual prototypes of compositions.

\textbf{Global Visual Prototype Codebook and Counterfactual Composition.} While the factor-specific features $\zv^s$ and $\zv^o$ provide controllable building blocks for compositions, we next refine the visual prototypes, beginning by constructing a global visual codebook:
\begin{equation}
\mathbf{H}=\{\hv_{m,n}^c\in\mathbb{R}^d\mid m=1,\dots,M;\,n=1,\dots,N\},
\end{equation}
where each $\hv_{m,n}^c$ encodes the interaction between state $s_m$ and object $o_n$, with one prototype per composition $c=(s_m,o_n)$. The codebook can be initialized either randomly or using CLIP features of seen pairs. To expose \emph{unseen} entries in $\mathbf{H}$ to visual evidence, we perform \emph{counterfactual recomposition} within each mini-batch by pairing factors across images (fixing one factor and intervening on the other) to synthesize plausible yet unseen labels. Specifically, let
\begin{equation}
\mathbf{z}^{\text{cf}}_{(i,j)}=\phi\!\big(\mathbf{z}^s_i,\mathbf{z}^o_j\big)\in\mathbb{R}^d,\qquad
c_{(i,j)}=(s_{m_i},o_{n_j}),   
\end{equation}
where $m_i$ and $n_j$ are the (seen) state/object indices of $\mathbf{x}_i$ and $\mathbf{x}_j$, respectively.
{To incorporate an additive compositional prior similar to semantic embeddings, while adapting to the visual manifold via non-linearity, we define the fusion function $\phi\!\big(\mathbf{z}^s_i,\mathbf{z}^o_j\big) = \sigma(\mathbf{M}_s \mathbf{z}_i^s + \mathbf{M}_o \mathbf{z}_j^o)$, where $\mathbf{M}_{s}$ and $\mathbf{M}_{o}$ are learnable projections and $\sigma$ denotes ReLU activation.}
These counterfactual composites will be used to update the corresponding entries in $\mathbf{H}$ even when $c_{(i,j)}$ is unseen, providing a direct pathway for visual evidence to accrue to prototypes beyond the training pairs.

\begin{algorithm}[ht]

\SetAlgoLined
\DontPrintSemicolon
\caption{Dynamic Visual-Prototype Refinement (Local Graph)}
\label{alg:local_graph}

\NonHangingInput{Input}{ Batch Factors $\mathbf{Z}_s=\{\zv^s_i\}, \mathbf{Z}_o=\{\zv^o_i\}$, Counterfactuals $\mathbf{Z}_{\text{cf}}=\{\zv^{\text{cf}}_{(i,j)}\}$, 
Global Codebook $\mathbf{H}$,  
Candidate Labels $\mathcal{C}_{\text{cand}}$,  
Momentum $\lambda$
}
\KwOut{Refined Global Codebook $\mathbf{H}$}

\tcp{1. Local Graph Construction}
Identify relevant prototypes: $\mathbf{H}_{\text{cand}} = \{\hv^c \in \mathbf{H} \mid c \in \mathcal{C}_{\text{cand}}\}$\;
Construct Node Features: $\mathbf{X}_{\text{local}} = [\mathbf{H}_{\text{cand}}; \mathbf{Z}_s; \mathbf{Z}_o; \mathbf{Z}_{\text{cf}}]$ \tcp*{(Eq.14)}
Compute Adjacency Matrix $\mathbf{W}$ using posteriors $p(s|\xv), p(o|\xv)$ \tcp*{(Eq.15)}

\tcp{2. Aggregation (Message Passing)}
Compute normalized aggregation weights $\mathbf{D}_P^{-1}\mathbf{W}_{P,:}$\;
Refine prototypes: \tcp*{(Eq.16)}
$\hat{\mathbf{H}}_{\text{cand}} \leftarrow \mathcal{A}(\mathbf{H}_{\text{cand}}; \mathbf{Z}_s, \mathbf{Z}_o, \mathbf{Z}_{\text{cf}}) = \mathbf{D}_P^{-1}\mathbf{W}_{P,:}\mathbf{X}_{\text{local}}$ 

\tcp{3. Momentum Update}
\For{each composition $c \in \mathcal{C}_{\text{cand}}$ \tcp*{(Eq.17-18)}} 
{ Update global entry: $\hv^c \leftarrow \lambda \hv^c + (1 - \lambda) \hat{\hv}^c$ }

\Return $\mathbf{H}$

\end{algorithm}

\subsection{Dynamic Visual-Prototype Refinement}
With the global codebook $\mathbf{H}$ initialized and counterfactual composites $\mathbf{Z}_{\text{cf}}$ available, the remaining question is how to route instance-level factor evidence $(\mathbf{Z}_s,\mathbf{Z}_o,\mathbf{Z}_{\text{cf}})$ into the appropriate prototypes in $\mathbf{H}$ while respecting the target set $\mathcal{C}_{tgt}$. We address this by constructing a label-conditioned local graph for each mini-batch and aggregating messages via a normalized operator $\mathcal{A}$ to obtain refined prototypes $\widehat{\mathbf{H}}$.

\textbf{Label-Conditioned Mini-Batch Local Graph.} To inject visual evidence into \emph{unseen} compositions during training, we build a \emph{label-conditioned} local graph on each mini-batch \(\mathcal{B}\). Let
\begin{equation}
\mathcal{S}(\mathcal{B})=\{\,s_{m_i}\!:\,\xv_i\!\in\!\mathcal{B}\,\},\quad
\mathcal{O}(\mathcal{B})=\{\,o_{n_j}\!:\,\xv_j\!\in\!\mathcal{B}\,\}. 
\end{equation}
We then form the batch-dependent candidate set of feasible composition labels as
\begin{equation}
\mathcal{C}_{\text{cand}}
\;=\;
\big(\,\mathcal{S}(\mathcal{B})\times \mathcal{O}(\mathcal{B})\,\big)\;\cap\; \mathcal{C}_{tgt},
\end{equation}
i.e., all state–object pairs that can be composed from factors present in the current batch and are permitted by the test-time target set \(\mathcal{C}_{tgt}\). Hence \(\mathcal{C}_{\text{cand}}\) may include both \emph{seen} and \emph{unseen} labels; unseen labels have no images in the batch but remain valid graph targets.

\emph{Nodes and features.}
In each mini-batch, we construct a \emph{local subgraph} containing only the prototypes relevant to the current candidate set $\mathcal{C}_{\text{cand}}$. 
Let $d$ be the feature dimension and $|\cdot|$ denote set cardinality. 
The node set of this local graph consists of: \textbf{(1) prototype nodes:} $\mathbf{H}_{\text{cand}} = \{\mathbf{h}^{c}_{m,n}\mid (s_m,o_n)\in\mathcal{C}_{\text{cand}}\}\in\mathbb{R}^{|\mathcal{C}_{\text{cand}}|\times d}$, a subset of the global codebook $\mathbf{H}$. \textbf{(2) factor nodes:} state features $\mathbf{Z}_s=\{\mathbf{z}^s_i\}_{i\in\mathcal{B}}$ and object features $\mathbf{Z}_o=\{\mathbf{z}^o_j\}_{j\in\mathcal{B}}$ extracted from current batch samples. \textbf{(3) counterfactual nodes:} $\mathbf{Z}_{\text{cf}}=\{\mathbf{z}^{\text{cf}}_{(i,j)}\}$ obtained by cross-pairing state and object factors within the batch. We stack all node features row-wise as
\begin{equation}
\mathbf{X}_{\text{local}}
=\big[\mathbf{H}_{\text{cand}};\mathbf{Z}_s;\mathbf{Z}_o;\mathbf{Z}_{\text{cf}}\big]
\in\mathbb{R}^{N_{\text{local}}\times d},
\end{equation}
where $N_{\text{local}}=|\mathcal{C}_{\text{cand}}|+|\mathbf{Z}_s|+|\mathbf{Z}_o|+|\mathbf{Z}_{\text{cf}}|$. 
Only the prototypes in $\mathbf{H}_{\text{cand}}$ are updated during the current iteration, while the rest of $\mathbf{H}$ remain unchanged and serve as a persistent global memory.
{In implementation, $\mathbf{H}_{\text{cand}}$ is realized as an index view over the global codebook, so refinement does not allocate additional copies of visual prototypes; only their entries in $\mathbf{H}$ are updated in place.}
The global codebook $\mathbf{H}$ thus acts as a memory that accumulates progressively refined prototypes across mini-batches, while each local graph only updates the relevant subset $\mathbf{H}_{\text{cand}}$.

\emph{Edges.} Connectivity is determined \emph{exclusively} by label feasibility rather than geometric proximity in feature space. We only allow edges that deliver messages into prototype nodes \(\mathbf{h}^{c}_{m,n}\) whose label pairs \((s_m,o_n)\) belong to the candidate set \(\mathcal{C}_{\text{cand}}\).
Let \(\mathbf{I}\{\cdot\}\) denote the indicator function, and let \(p(s_m\mid \xv)\) and \(p(o_n\mid \xv)\) be the factor posteriors introduced earlier.
Each edge weight \(w(v\!\to\!u)\) quantifies the contribution from a source node \(v\) to a destination prototype node \(u\), defined as:

\begin{equation}
\resizebox{0.9\hsize}{!}{$
\begin{aligned}
w\!\left(\mathbf{z}^s_i \!\to\! \mathbf{h}^{c}_{m,n}\right)
&=\frac{\mathbf{I}\{(s_m,o_n)\in \mathcal{C}_{\text{cand}}\}}{|\mathcal{O}_{\text{cand}}(m)|} \cdot p(s_m\mid \xv_i),\\[-1mm]
\mathcal{O}_{\text{cand}}(m)&=\{o_n:(s_m,o_n)\in \mathcal{C}_{\text{cand}}\};\\[2mm]
w\!\left(\mathbf{z}^o_j \!\to\! \mathbf{h}^{c}_{m,n}\right)
&=\frac{\mathbf{I}\{(s_m,o_n)\in \mathcal{C}_{\text{cand}}\}}{|\mathcal{S}_{\text{cand}}(n)|} \cdot p(o_n\mid \xv_j),\\
\mathcal{S}_{\text{cand}}(n)&=\{s_m:(s_m,o_n)\in \mathcal{C}_{\text{cand}}\};\\[2mm]
w\!\left(\mathbf{z}^{\text{cf}}_{(i,j)} \!\to\! \mathbf{h}^{c}_{m,n}\right)
&=\mathbf{I}\{(m,n)=(m_i,n_j)\} \cdot p(s_{m_i}\!\mid \xv_i)\,p(o_{n_j}\!\mid \xv_j).
\end{aligned}
$}
\end{equation}

These label-conditioned edges define a sparse, directed adjacency matrix
\(\mathbf{W}\in\mathbb{R}^{N_{\text{local}}\times N_{\text{local}}}\),
where each entry \(\mathbf{W}_{u,v}=w(v\!\to\!u)\) stores the message strength from node \(v\) to node \(u\).
Only prototype nodes (the first \(|\mathcal{C}_{\text{cand}}|\) destination rows) receive nonzero incoming edges.
Let \(\mathbf{1}\) be an all-ones vector and define the row-degree matrix as
\(\mathbf{D}=\operatorname{diag}(\mathbf{W}\,\mathbf{1})\).
Because connectivity depends solely on label feasibility, every unseen prototype
\(\mathbf{h}^{c}_{m,n}\in\mathbf{H}_{\text{cand}}\)
receives nonzero gradients from seen factors and counterfactual pairs, maintaining update paths without unseen images.

\emph{Explicit aggregation function \(\mathcal{A}(\cdot)\).}
Given the adjacency structure above, we perform message aggregation into prototype nodes using a degree-normalized weighted average.
Let \(\mathbf{W}_{P,:}\in\mathbb{R}^{|\mathcal{C}_{\text{cand}}|\times N_{\text{local}}}\) denote the submatrix of \(\mathbf{W}\) corresponding to prototype-destination rows (i.e., rows associated with \(\mathbf{H}_{\text{cand}}\)).
We define the sub-degree matrix \(\mathbf{D}_P=\operatorname{diag}(\mathbf{W}_{P,:}\mathbf{1})\).
Let \(\mathbf{X}_{\text{local}}\in\mathbb{R}^{N_{\text{local}}\times d}\) collect all local nodes.
The explicit aggregation operation is:
\begin{equation}
\label{eq:aggregation}
\begin{aligned}
&\widehat{\mathbf{H}}_{\text{cand}}
= \mathcal{A}\big(\mathbf{H}_{\text{cand}}; \mathbf{Z}_s,\mathbf{Z}_o,\mathbf{Z}_{\text{cf}}\big)
= \mathbf{D}_P^{-1}\,\mathbf{W}_{P,:}\,\mathbf{X}_{\text{local}}, \\
&\widehat{\mathbf{H}}_{\text{cand}} \in \mathbb{R}^{|\mathcal{C}_{\text{cand}}|\times d}, \quad
\mathbf{D}_P \in \mathbb{R}^{|\mathcal{C}_{\text{cand}}|\times|\mathcal{C}_{\text{cand}}|}.
\end{aligned}
\end{equation}

Here, $\mathbf{D}_P^{-1}\,\mathbf{W}_{P,:}$ acts as a row-normalized attention operator that gathers messages from connected sources into each prototype.
Dimensionally, \(\mathbf{W}_{P,:}\mathbf{X}_{\text{local}}\) maps
\((|\mathcal{C}_{\text{cand}}|\times N_{\text{local}})\) by \((N_{\text{local}}\times d)\)
into updated prototype embeddings \((|\mathcal{C}_{\text{cand}}|\times d)\),
and \(\mathbf{D}_P^{-1}\) performs per-prototype degree normalization.

{Crucially, the aggregation process in Eq.\eqref{eq:aggregation} establishes a fully differentiable pathway for unseen compositions. Although unseen prototypes do not correspond to ground-truth images in the current mini-batch, they actively aggregate features from the disentangled seen factors ($\mathbf{Z}_s, \mathbf{Z}_o$) and counterfactual pairs. During the optimization of $\mathcal{L}_c^h$, these refined unseen prototypes serve as negative classes, receiving gradient updates that back-propagate to the shared disentanglers. The detailed dynamic refinement procedure is described in Algorithm~\ref{alg:local_graph}.}

As illustrated in Fig.~\ref{fig:local-graph}, edges are established based on semantic feasibility between disentangled factors and compositional prototypes. For example, if the mini-batch contains features corresponding to the state ``old'' and the object ``tiger,'' these nodes connect to the prototype of the unseen composition ``old tiger,'' enabling contextualized updates even without unseen images. Consequently, only a subset of the global prototypes is updated at each iteration, while the remainder in \(\mathbf{H}\) remains unchanged and serves as persistent memory. After aggregation, we apply a momentum update and write the refined \(\widehat{\mathbf{H}}_{\text{cand}}\) back to the corresponding entries in the global codebook \(\mathbf{H}\).

\textbf{Momentum Update and Regularization.}
To stabilize temporal dynamics while allowing adaptation, we apply a momentum write-back \emph{only} to the prototypes involved in the current local graph, {which can be viewed as an exponential moving average (EMA) between the historical global prototypes and the current batch-level aggregated ones}:
\begin{equation}
\mathbf{H}_{\text{cand}}
\;\leftarrow\;
\lambda\,\mathbf{H}_{\text{cand}}
+(1-\lambda)\,\widehat{\mathbf{H}}_{\text{cand}},
\qquad \lambda\in[0,1],
\end{equation}
and then commit the updated rows back to the global codebook \(\mathbf{H}\) at the corresponding indices of \(\mathcal{C}_{\text{cand}}\). {In this way, each mini-batch contributes only a small incremental correction scaled by $(1-\lambda)$ on top of the accumulated prototypes $\lambda\,\mathbf{H}_{\text{cand}}$, which damps the influence of any single noisy or imbalanced mini-batch and smooths prototype evolution over training.} We denote the global codebook after this write-back by $\hat{\mathbf{H}}$:
\begin{equation}
\begin{aligned}
\hat{\mathbf{H}}
&\!=\!\big\{\hat{\hv}^{c}_{m,n}\in\mathbb{R}^{d}\big\}_{m,n},\\
\hat{\hv}^{c}_{m,n}&\!=\!
\begin{cases}
\lambda\,\hv^{c}_{m,n}+(1-\lambda)\,\hat{\hv}^{c}_{m,n}, &(s_m,o_n)\!\in\!\mathcal{C}_{\text{cand}},\\
\hv^{c}_{m,n}, & \text{otherwise},
\end{cases}
\end{aligned}
\end{equation}
where we optionally re-normalize each updated prototype to unit length. Entries not in $\mathcal{C}_{\text{cand}}$ remain unchanged {and thus act as stable anchors for compositions that do not appear in the current mini-batch. In practice, we tune $\lambda$ on a validation split in the range $[0, 1]$ (Tab.~\ref{tab:ablation_update} and Fig.~\ref{fig:Sensitivity_update}) and select a relatively high value. This choice makes the update dominated by historical prototypes, reducing the impact of early noisy predictions on the global codebook while still allowing gradual adaptation through batch-level local graphs.}

\noindent\emph{Prototype-to-image alignment.}
To regularize the refined prototypes and encourage semantic consistency with global visual evidence, we align them with the compositional visual features \(\zv_i^c\) (Eq.~\ref{eq:imgfeat}) via a softmax:
\begin{equation}
\begin{aligned}
p(c_{m,n}^{h}\mid \xv_i;\hat{\mathbf{H}})&=\frac{\exp\!\big(\zv_i^c\cdot \hat{\hv}^{c}_{m,n}/\tau\big)}
{\sum_{m=1}^{M}\sum_{n=1}^{N}\exp\!\big(\zv_i^c\cdot \hat{\hv}^{c}_{m,n}/\tau\big)},\\
\mathcal{L}_c^h&=-\frac{1}{|\mathcal{T}_{tr}|}\sum_{i}\log p(c_{m,n}^h\mid \xv_i;\hat{\mathbf{H}}),
\end{aligned}
\end{equation}
This regularization ensures that the EMA-updated prototypes remain semantically meaningful and discriminative with respect to the global visual feature space. {It complements the momentum update and further enhances compositional generalization across both seen and unseen state-object pairs.}

\textbf{Computational complexity.}
The computational cost of the local graph module is dominated by interactions among the state and object factor nodes, the instantiated composition nodes, and the candidate prototype set $\mathbf{H}_{\text{cand}}$. Let $B$ be the batch size, and let $S_b$ and $O_b$ be the numbers of distinct states and objects in a mini-batch (so $S_b, O_b \le B$). We disentangle state and object features per image and aggregate them into $S_b$ state factor nodes and $O_b$ object factor nodes. The local graph activates only compositions $\mathcal{C}_{\text{cand}}$ and updates only the corresponding rows of $\mathbf{H}_{\text{cand}}$, with $|\mathcal{C}_{\text{cand}}|\le \min(S_bO_b,|\mathcal{C}_{\text{tgt}}|)$. For each $c=(s,o)\in\mathcal{C}_{\text{cand}}$, the module (i) aggregates normal and counterfactual instances to compute $z_{(s,o)}$ from the associated factor nodes and (ii) refines the corresponding prototype in $\mathbf{H}_{\text{cand}}$ (e.g., via EMA). Each step uses a constant number of $d$-dimensional vectors per label, so the prototype-side cost is $O(|\mathcal{C}_{\text{cand}}|\,d)$.

Therefore, the per-iteration complexity of the local graph module is $O(|\mathcal{C}_{\text{cand}}|\,d)$ on top of the backbone encoding cost $O(B\,d)$, and the overhead scales with the number of active compositions in the mini-batch rather than the full state-object Cartesian product. Empirically (Sec.~4.3, Tab.~\ref{tab:efficiency}), most memory and runtime overhead comes from the global codebook (Duplex without L.G.), while the local graph adds only marginal additional cost. The local graph is used only during training; at inference, we discard it and score with frozen prototypes, matching a standard CLIP-based prototype classifier.

\subsection{Training Objective and Inference}
\textbf{Training objective.}
During training, we define the overall loss as

\begin{equation}
\mathcal{L}=\mathcal{L}_{s}+\mathcal{L}_{o}+\mathcal{L}^{t}_{c}+\mathcal{L}^{h}_{c},
\end{equation}
where $\mathcal{L}^{t}_{c}$ aligns \emph{semantic} prototypes from text and $\mathcal{L}^{h}_{c}$ aligns \emph{visual} prototypes refined by the local graph. All features are $\ell_2$-normalized.

\textbf{Inference.}
Given a test image \(\xv\), we compute four sets of probabilities: \(p(c^{h}_{m,n}\!\mid\!\xv)\) (visual prototypes), \(p(c^{t}_{m,n}\!\mid\!\xv)\) (semantic prototypes), and \(p(s_{m}\mid\xv)\), \(p(o_{n}\mid\xv)\) (independent factors). We fuse them as
\begin{equation}
\begin{aligned}
\texttt{\large\textbf{S}}(c_{m,n}\mid \xv)=
p(c^{h}_{m,n}\mid\xv)&+p(c^{t}_{m,n}\mid\xv)\\
&+p(s_{m}\mid\xv)\cdot p(o_{n}\mid\xv),
\end{aligned}
\end{equation}
where the product term serves as a bias correction under an independence assumption. The final prediction is
\begin{equation}
\widetilde{c}=\underset{c_{m,n} \in \; \mathcal{C}_{tgt}}
{\arg \max}\;\texttt{\large\textbf{S}}(c_{m, n}\mid \xv).
\end{equation}
where \(\widetilde{c}\) denotes the predicted composition class.
{The complete training procedure, which integrates semantic prototype construction, state-object disentanglement, and dynamic visual-prototype refinement, is summarized in Algorithm~\ref{alg:duplex_framework}.}

\begin{algorithm}[ht]

\SetAlgoLined
\DontPrintSemicolon
\caption{Duplex Training and Inference}
\label{alg:duplex_framework}

\NonHangingInput{Input}{
    Mini-batch images $\mathcal{B} = \{\xv_i\}_{i=1}^{N_{bs}},$,
    Global visual codebook $\mathbf{H}$, text/image encoders $E_{\text{txt}}, E_{\text{img}}$,
    Disentanglers $D^s, D^o$, prompts $\mathbf{P}^c,\mathbf{P}^s,\mathbf{P}^o$
}
\KwOut{Predicted composition labels $\{\tilde{c}_i\}_{i=1}^{N_{bs}}$}

\tcp{1. Semantic Prototypes and Primitives}
Semantic prototypes:
$\tv^c \leftarrow E_{\text{txt}}(\mathbf{P}^c)$ \tcp*{(Eq.2)}
State/Object Semantic Primitives:
$\tv^s, \tv^o \leftarrow E_{\text{txt}}(\mathbf{P}^s), E_{\text{txt}}(\mathbf{P}^o) $ \tcp*{(Eq.7)}

\tcp{2. Image encoding and disentangled visual factors}
Global visual feature: $\zv^{\text{cls}}_i \leftarrow E_{\text{img}}(\xv_i)$ \tcp*{(Eq.3)}
Disentangled visual factors: $\zv^s_i,\zv^o_i \leftarrow D^s(\zv^{\text{cls}}_i),D^o(\zv^{\text{cls}}_i)$ \tcp*{(Eq.5)}

\tcp{3. Dynamic Visual-Prototype Refinement}
\eIf{Training Phase}{
    Generate $\mathbf{Z}_{\text{cf}}$ from batch factors \tcp*{(Eq.11)}
    Construct Local Graph and Update $\mathbf{H}_{\text{cand}}$ \;
    $\hat{\mathbf{H}} \leftarrow \textbf{Algorithm \ref{alg:local_graph}}\;(\mathbf{H}, \mathbf{Z}_s, \mathbf{Z}_o, \mathbf{Z}_{\text{cf}}, \mathcal{C}_{\text{cand}})$\;
}{
    Use Global Codebook $\mathbf{H}$\; 
    (No graph update during inference)\;
}

\tcp{4. Prediction and training objective}
For all $\xv_i \in \mathcal{B}$, 
using current $\mathbf{H}$, $\tv^c,\tv^s,\tv^o$ compute Semantic/Visual Composition and State/Object Posteriors: 
$p(c^t_{m,n}\mid \xv_i)$, $p(c^h_{m,n}\mid \xv_i;\mathbf{H})$, $p(s_m\mid \xv_i)$, $p(o_n\mid \xv_i)$; Fuse them to obtain scores
$S(c_{m,n}\mid\xv_i)$, then predict
$\tilde{c}_i$. \tcp*{(Eq.21-22)}

\Return $\{\tilde{c}_i\}_{i=1}^{N_{bs}}$
\end{algorithm}

\section{Experimental Evaluation}
\label{exp}

\subsection{Experimental setup}
\textbf{Datasets.}
We experiment with three real-world CZSL benchmarks: MIT-States, UT-Zappos, and CGQA. {MIT-States and CGQA are typically characterized as attribute--object datasets, whereas UT-Zappos contains fine-grained attributes such as material or style. To ensure terminological consistency across these benchmarks, we follow the CZSL literature and adopt the term ``state'' as a unified abstraction for all visual primitives that modify objects (e.g., color, texture, or material). Formally, these primitives are mapped into a unified state set $\mathcal{S}$.} We summarize detailed statistics in Tab.~\ref{tab:dataset_statistics}.

{\textbf{Evaluation Metrics.}
Following established protocols in the CZSL literature~\cite{naeem2021learning, Purushwalkam_2019_ICCV, invariant_2022ECCV, csp2023, mancini2021open}, we evaluate performance using four standard metrics: Area Under the Curve (AUC), the best Harmonic Mean (HM), and the corresponding Top-1 accuracies on Seen (S) and Unseen (U) sets.
In both closed-world ~\cite{Purushwalkam_2019_ICCV} and open-world~\cite{mancini2021open} CZSL settings, the Harmonic Mean serves as the unified measure of the trade-off between seen and unseen composition recognition, calculated as:
\begin{equation}
    \text{HM} = 2 * (S * U) / (S + U)
    \label{eq:hm}
\end{equation}
where $S$ and $U$ denote the accuracies on seen and unseen validation/test splits, respectively. Furthermore, regardless of the specific setting, we prioritize AUC as the primary metric for evaluation and model selection. AUC provides a holistic assessment of the model's capability across the entire range of operating points, avoiding the potential bias of a single peak HM value.}

\begin{table}[!h]
\setlength{\tabcolsep}{1.5pt}
\renewcommand{\arraystretch}{0.9}
\centering
\caption{Dataset statistics for UT-Zappos, MIT-States, and CGQA.}
\resizebox{\linewidth}{!}{%
\begin{tabular}{@{}l ccc| cc| ccc| ccc@{}}
\toprule
& & & & \multicolumn{2}{c|}{$\textit{Train}$} & \multicolumn{3}{c|}{$\textit{Val}$} & \multicolumn{3}{c}{$\textit{Test}$} \\
\textit{Datasets} & $\mathcal{|S|}$ & $\mathcal{|O|}$ & $\mathcal{|S| \times |O|}$ & $|\mathcal{C}_s| $ & $|\mathcal{X}_s| $ & $|\mathcal{C}_s|$ & $|\mathcal{C}_u|$ & $\textit{Imgs}$ & $|\mathcal{C}_s|$ & $|\mathcal{C}_u|$ & $\textit{Imgs}$ \\
\midrule
UT-Zappos~\cite{UT-Zappos} & 16 & 12 & 192 & 83 & 22,998 & 15 & 15 & 3,214 & 18 & 18 & 2,914 \\
MIT-States~\cite{MIT-States} & 115 & 245 & 28,175 & 1,262 & 30,338 & 300 & 300 & 10,420 & 400 & 400 & 19,191 \\
CGQA~\cite{naeem2021learning} & 413 & 674 & 278,362 & 5,592 & 26,920 & 1,252 & 1,040 & 7,280 & 888 & 923 & 5,098 \\
\bottomrule
\end{tabular}%
}
\label{tab:dataset_statistics}
\end{table}

\begin{table}[!h]
\setlength{\tabcolsep}{2pt}
\renewcommand{\arraystretch}{0.8}
    \centering
    \caption{Hyperparameters for different datasets.}
    \resizebox{0.96\linewidth}{!}{
    \begin{tabular}{cccc}
    \toprule
    Hyperparameter & MIT-States & UT-Zappos & CGQA \\
    \midrule
    Learning rate  &  $10^{-4}$& $5 \times 10^{-4}$ &$1.5 \times 10^{-5}$ \\
       Batch size &  64 & 64 & 32 \\
       Number of epochs &  15 & 20 & 20\\
       graph node dimension&  768& 768 & 768\\
       state\&object disentangler dimension&  768& 768 & 768\\
    \bottomrule
    \end{tabular}}
    \label{tab:Hyperparameters}
\end{table}

\begin{figure}[pos=th]
\centering
\includegraphics[width=1.0\linewidth]{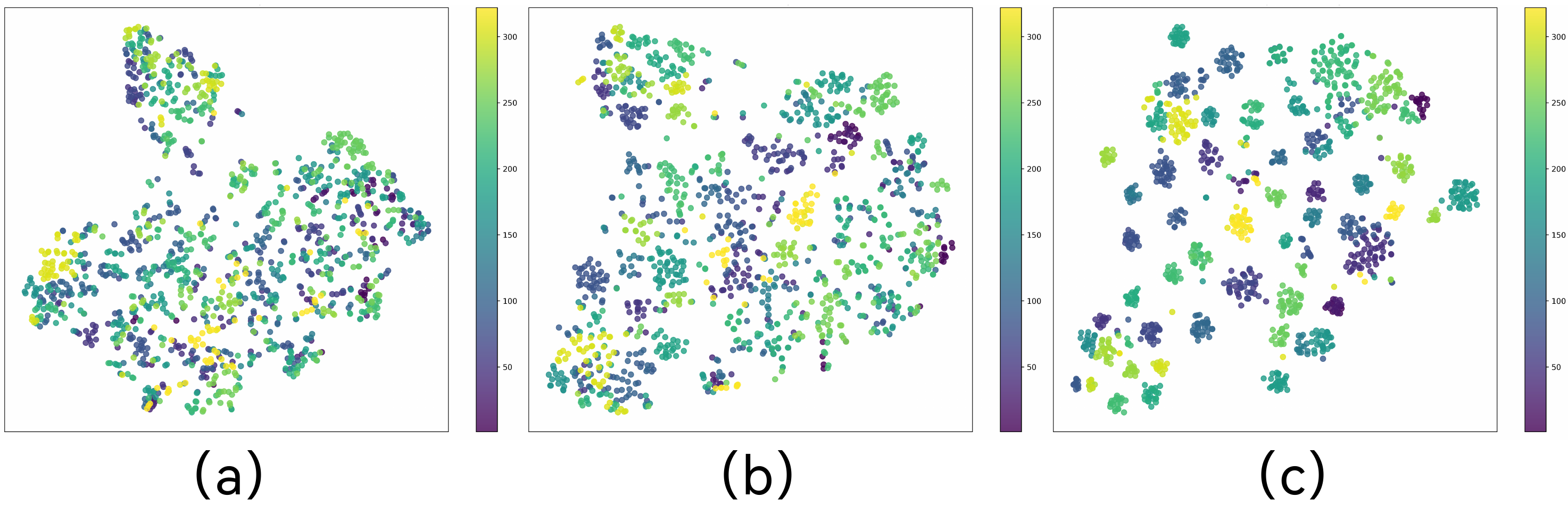}
\caption{Embedding visualization using the first 300 classes from MIT-States for clustering, with colors indicating different categories. We show (a) semantic-prototype embeddings, (b) visual-prototype embeddings, and (c) combined semantic and visual prototype embeddings.}
\label{fig:text_vison_embeddingshow}
\end{figure}

\textbf{Implementation Details.} \emph{Duplex} is implemented with a pretrained CLIP ViT-L/14 model.  We train and evaluate on a single NVIDIA A800 GPU using the seen split. During testing in the open-world setting, we apply post-training calibration~\cite{csp2023} to filter infeasible compositions. Tab.~\ref{tab:Hyperparameters} lists the dataset-specific hyperparameters, which are selected based on validation performance. For the remaining hyperparameters, we initialize all three prefixes with CLIP's pretrained prompt ``a photo of.'' For the adapter inserted into the image encoder, the bottleneck dimension $r$ is set to 64 and the dropout rate is set to 0.1.  During training, we use the Adam optimizer and decay the learning rate of all trainable parameters by a factor of 0.5 every 5 epochs.

\begin{table*}[!h]
\begin{center}
\caption{ Quantitative results in a \textbf{CW} setting on three benchmarks.  
All methods with the prefix \textit{w/ CLIP} utilize CLIP ViT-L/14 as backbone.  
$\pm$ denotes the standard error of the mean over five random model initializations.  
\textbf{Bold numbers} indicate the best performance and \underline{underlined numbers} indicate the second best.}

\setlength{\tabcolsep}{3pt}
\renewcommand{\arraystretch}{0.9}

\resizebox{0.9\linewidth}{!}{
\begin{tabular}{
c|c| ccc c| ccc c|ccc c }
\toprule
 &
\multicolumn{1}{l|}{\multirow{2}{*}{{Method}}}
& \multicolumn{4}{c|}{ {MIT-States }} 
& \multicolumn{4}{c|}{ {UT-Zappos }} 
& \multicolumn{4}{c}{ {CGQA }} \\
 
&\multicolumn{1}{l|}{}
& S & U & HM & AUC 
&S & U & HM & AUC 
& S & U & HM & AUC \\
 \midrule
\multirow{8}{*}{\rotatebox{90}{\centering w/o CLIP}}&\multicolumn{1}{l|}{CompCos~\cite{mancini2021open} }&25.3& 24.6& 16.4& 4.5& 59.8& 62.5& 43.1& 28.1& 28.1& 11.2& 12.4& 2.6\\
&\multicolumn{1}{l|}{CGE~\cite{naeem2021learning}} &28.7 &25.3& 17.2& 5.1& 56.8& 63.6& 41.2& 26.4& 28.1& 10.1& 11.4& 2.3\\
&\multicolumn{1}{l|}{Co-CGE~\cite{Graph-TPAMI}}  &27.8& 25.2& 17.5 &5.1 &58.2 &63.3 &44.1 &29.1 &29.3 &11.9 &12.7 &2.8\\
&\multicolumn{1}{l|}{SCEN~\cite{Li_2022_CVPR}} &29.9 &25.2 &18.4 &5.3 &63.5 &63.1 &47.8 &32.0 &28.9 &12.1 &12.4 &2.9\\
&\multicolumn{1}{l|}{CVGAE~\cite{2022MMCVGAE}} &28.5 &25.5 &18.2 &5.3 &65.0 &62.4 &49.8 &34.6 &28.2 &11.9 &13.9 &2.8 \\
&\multicolumn{1}{l|}{CANet~\cite{CANET_2023CVPR}} &29.0 &26.2 &17.9 &5.4 &61.0 &66.3 &47.3 &33.1 &30.0 &13.2 &14.5 &3.3 \\
&\multicolumn{1}{l|}{CAPE~\cite{Khan_2023_WACV}} & 30.5 &26.2 &19.1 &5.8 &60.4 &67.4 &45.5 &31.3 &32.9 &15.6 &16.3 &4.2 \\
&\multicolumn{1}{l|}{ADE~\cite{hao2023ade}} & —& —& —& —& 63.0& 64.3& 51.1& 35.1& 35.0& 17.7& 18.0& 5.2 \\
&\multicolumn{1}{l|}{LVAR-CZSL~\cite{2024TLVAR-CZSL}} &33.3&26.7&19.5&6.4&63.8&67.6&53.1&37.6&36.0&18.2&18.5&5.4 \\

\midrule
 \multirow{10}{*}{\rotatebox{90}{\centering w/ CLIP}}&\multicolumn{1}{l|}{CLIP~\cite{pmlr-v139-radford21a}} &  30.2 & 46.0 & 26.1 & 11.0 & 15.8 & 49.1 & 15.6 & 5.0 & 7.5 & 25.0 & 8.6 & 1.4 \\
 &\multicolumn{1}{l|}{CoOp~\cite{zhou2022coop}} &  34.4 & 47.6 & 29.8 & 13.5 & 52.1 & 49.3 & 34.6 & 18.8 & 20.5 & 26.8 & 17.1 & 4.4 \\
 &\multicolumn{1}{l|}{CSP~\cite{csp2023}} &  46.6 & 49.9 & 36.3 & 19.4 & 64.2 & 66.2 & 46.6 & 33.0 & 28.8 & 26.8 & 20.5 & 6.2 \\
 &\multicolumn{1}{l|}{GIPCOL~\cite{GIPCOL}} &  48.5 &49.6 &36.6 &19.9 &65.0 &68.5 &48.8 &36.2 &31.9 &28.4 &22.5 &7.1 \\
 &\multicolumn{1}{l|}{DFSP(i2t)~\cite{Lu_2023_CVPR}} &  47.4 & 52.4 & 37.2 & 20.7 & 64.2 & 66.4 & 45.1 & 32.1 & 35.6 & 29.3 & 24.3 & 8.7 \\
 &\multicolumn{1}{l|}{DFSP(BiF)~\cite{Lu_2023_CVPR}} &  47.1 & 52.8 & 37.7 & 20.8 & 63.3 & 69.2 & 47.1 & 33.5 & 36.5 & 32.0 & 26.2 & 9.9 \\
 &\multicolumn{1}{l|}{DFSP(t2i)~\cite{Lu_2023_CVPR}} &  46.9 & 52.0 & 37.3 & 20.6 & 66.7 & 71.7 & 47.2 & 36.0 & 38.2 & 32.0 & 27.1 & 10.5 \\
 &\multicolumn{1}{l|}{Troika~\cite{Huang2024Troika}} &  49.0 & {53.0}  & {39.3} & 22.1 & {66.8}& 73.8 & {54.6} & {41.7} & {41.0} & {35.7} & {29.4} & {12.4} \\
 &\multicolumn{1}{l|}{CDS-CZSL~\cite{2024CVPR-CDSCZSL}} &  \underline{50.3} & 52.9  & 39.2 & {22.4} & 63.9& {74.8} & 52.7 & 39.5 & 38.3 & 34.2 & 28.1 & 11.1 \\
 &\multicolumn{1}{l|}{LOGICZSL\cite{2025LOGICZSL}}
 &\textbf{50.8} &\underline{53.9} &\underline{40.5} &\underline{23.4} &\underline{69.6} &\underline{74.9} &\underline{57.8} &\underline{45.8} &\textbf{44.4} &\textbf{39.4} &\textbf{33.3} &\textbf{15.3}\\

 \cellcolor{white}&\multicolumn{1}{l|}{\textbf{\emph{Duplex}(ours)}} 
 & ${\text{49.7} {\scriptscriptstyle \pm 0.3}}$ 
 & $\textbf{55.6} {\scriptscriptstyle \pm 0.4}$  
 & $\textbf{40.9}{\scriptscriptstyle \pm 0.2}$ 
 & $\textbf{23.7}{\scriptscriptstyle \pm 0.4}$ 
 
 & $\textbf{70.5}{\scriptscriptstyle \pm 0.5}$ 
 & $\textbf{75.6}{\scriptscriptstyle \pm 0.9}$ 
 & $\textbf{58.2}{\scriptscriptstyle \pm 0.9}$ 
 & $\textbf{46.2}{\scriptscriptstyle \pm 0.7}$
 
 & $\underline{\text{41.1}}{\scriptscriptstyle \pm 0.2}$ 
 & $\underline{\text{36.2}}{\scriptscriptstyle \pm 0.4}$ 
 & $\underline{\text{30.1}}{\scriptscriptstyle \pm 0.2}$ 
 & $\underline{\text{13.2}}{\scriptscriptstyle \pm 0.4}$ \\
 \bottomrule

\end{tabular}}

 \label{tab:main_result_close}
\end{center}
\end{table*}

 \begin{table*}[!hbt]

\begin{center}
\caption{ Quantitative results in a \textbf{OW} setting on three benchmarks.  
All methods with the prefix \textit{w/ CLIP} utilize CLIP ViT-L/14 as backbone.  
$\pm$ denotes the standard error of the mean over five random model initializations.  
\textbf{Bold numbers} indicate the best performance and \underline{underlined numbers} indicate the second best.}

\setlength{\tabcolsep}{3pt}
\renewcommand{\arraystretch}{0.9}

\resizebox{0.9\linewidth}{!}{
\begin{tabular}{
c|c|ccc c|ccc c|ccc c }
\toprule
 &\multicolumn{1}{l|}{\multirow{2}{*}{{Method}}}
& \multicolumn{4}{c|}{ {MIT-States }} 
& \multicolumn{4}{c|}{ {UT-Zappos }} 
& \multicolumn{4}{c}{ {CGQA }} \\
 
&\multicolumn{1}{l|}{}
& S & U & HM & AUC 
& S & U & HM & AUC 
& S & U & HM & AUC  \\
\midrule

\multirow{7}{*}{\rotatebox{90}{\centering w/o CLIP}}&\multicolumn{1}{l|}{CompCos~\cite{mancini2021open} }&
25.4& 10.0& 8.9& 1.6& 59.3& 46.8& 36.9& 21.3& 28.4& 1.8& 2.8& 0.4\\

&\multicolumn{1}{l|}{CGE~\cite{naeem2021learning}}&
 29.6& 4.0& 4.9& 0.7& 58.8& 46.5& 38.0& 21.5& 28.3& 1.3& 2.2& 0.3\\

&\multicolumn{1}{l|}{Co-CGE~\cite{Graph-TPAMI}}&
 26.4& 10.4& 10.1& 2.0& 60.1& 44.3& 38.1& 21.3& 28.7& 1.6& 2.6& 0.4\\

&\multicolumn{1}{l|}{KG-SP~\cite{Karthik_2022_CVPR} }&
 28.4& 7.5& 7.4& 1.3& 61.8& 52.1& 42.3& 26.5& 31.5& 2.9& 4.7& 0.8\\

&\multicolumn{1}{l|}{CVGAE~\cite{2022MMCVGAE}}&
27.3& 9.9& 10.0& 1.8& 58.6& 48.4& 41.7& 22.2& 26.6& 2.9& 6.4& 0.7\\

&\multicolumn{1}{l|}{ADE~\cite{hao2023ade}}& 
 — & — & — & — & 62.4 & 50.7 & 44.8 & 27.1 & 35.1 & 4.8 & 7.6 & 1.4 \\

 &\multicolumn{1}{l|}{DRANet~\cite{Li_2023_ICCV}}& 
 29.8 &7.8 &7.9 &1.5 &65.1 &54.3 &44.0 &28.8 &31.3 &3.9 &6.0 &1.1 \\

 \midrule
 \multirow{10}{*}{\rotatebox{90}{\centering w/ CLIP}}
 &\multicolumn{1}{l|}{CLIP~\cite{pmlr-v139-radford21a}} & 30.1 & 14.3 & 12.8 & 3.0 & 15.7 & 20.6 & 11.2 & 2.2 & 7.5 & 4.6 & 4.0 & 0.3 \\
 &\multicolumn{1}{l|}{CoOp~\cite{zhou2022coop}} & 34.6 & 9.3 & 12.3 & 2.8 & 52.1 & 31.5 & 28.9 & 13.2 & 21.0 & 4.6 & 5.5 & 0.7 \\
 &\multicolumn{1}{l|}{CSP~\cite{csp2023}} & 46.3 & 15.7 & 17.4 & 5.7 & 64.1 & 44.1 & 38.9 & 22.7 & 28.7 & 5.2 & 6.9 & 1.2 \\
 &\multicolumn{1}{l|}{GIPCOL~\cite{GIPCOL}} & 48.5 &16.0 &17.9 &6.3 &65.0 &45.0 &40.1 &23.5 &31.6 &5.5 &7.3 &1.3 \\
 &\multicolumn{1}{l|}{DFSP(i2t)\cite{Lu_2023_CVPR}} & 47.2 & 18.2 & 19.1 & 6.7 & 64.3 & 53.8 & 41.2 & 26.4 & 35.6 & 6.5 & 9.0 & 2.0 \\
 &\multicolumn{1}{l|}{DFSP(BiF)~\cite{Lu_2023_CVPR}}& 47.1 & 18.1 & 19.2 & 6.7 & 63.5 & 57.2 & 42.7 & 27.6 & 36.4 & 7.6 & 10.6 & 2.4 \\
 &\multicolumn{1}{l|}{DFSP(t2i)\cite{Lu_2023_CVPR}} & 47.5 & 18.5 & 19.3 & 6.8 & {66.8} & 60.0 & 44.0 & 30.3 & 38.3 & 7.2 & 10.4 & 2.4 \\
 &\multicolumn{1}{l|}{Troika~\cite{Huang2024Troika}} & 48.8 & 18.7 & 20.1 & 7.2 & 66.4 & 61.2 & 47.8 & {33.0} & {40.8} & {7.9} & 10.9 & {2.7} \\
 &\multicolumn{1}{l|}{CDS-CZSL~\cite{2024CVPR-CDSCZSL}} & {49.4} & \underline{21.8}  & {22.1} & {8.5} & 64.7& {61.3} & {48.2} & 32.3 & 37.6 & {8.2} & {11.6} & {2.7} \\
  &\multicolumn{1}{l|}{LOGICZSL\cite{2025LOGICZSL}}
  &\textbf{50.7} &21.4 &\underline{22.4} &\underline{8.7} &\underline{69.6} &\underline{63.7} &\underline{50.8} &\underline{36.2} &\textbf{43.7} &\underline{9.3} &\textbf{12.6}&\underline{3.4}\\
 
  \cellcolor{white}&\multicolumn{1}{l|}{\textbf{\emph{Duplex}(ours)}} & $\underline{\text{50.6}{\scriptscriptstyle \pm 0.2}}$ 
  & $\textbf{22.0}{\scriptscriptstyle \pm 0.2}$ 
  & $\textbf{22.9}{\scriptscriptstyle \pm 0.1}$ 
  & $\textbf{9.0}{\scriptscriptstyle \pm 0.1}$
  
  & $\textbf{70.2}{\scriptscriptstyle \pm 0.2}$ 
  & $\textbf{64.8}{\scriptscriptstyle \pm 0.4}$ 
  & $\textbf{51.8}{\scriptscriptstyle \pm 0.3}$ 
  & $\textbf{37.3}{\scriptscriptstyle \pm 0.5}$
  
  & $\underline{\text{41.6}{\scriptscriptstyle \pm 0.3}}$ 
  & $\textbf{9.7}{\scriptscriptstyle \pm 0.2}$ 
  & $\underline{\text{12.5}{\scriptscriptstyle \pm 0.1}}$ 
  & $\textbf{3.4}{\scriptscriptstyle \pm 0.1}$ \\
 \bottomrule
\end{tabular}}
 \label{tab:main_result_open}
\end{center}
\end{table*}

\subsection{Compared Methods}
In this section, we compare our proposed \emph{Duplex} with CLIP-based baselines, including CLIP~\cite{pmlr-v139-radford21a}, CoOp~\cite{zhou2022coop}, CSP~\cite{csp2023}, GIPCOL~\cite{GIPCOL}, DFSP~\cite{Lu_2023_CVPR}, Troika~\cite{Huang2024Troika}, CDS-CZSL~\cite{2024CVPR-CDSCZSL}, and LOGICZSL~\cite{2025LOGICZSL}, under both CW and OW CZSL settings.

Across benchmarks, \emph{Duplex} delivers consistently strong results in the CW setting (Tab.~\ref{tab:main_result_close}), with clear gains on MIT-States and UT-Zappos and competitive performance on CGQA.
The latter shows a smaller margin to the strongest baseline, which we attribute to CGQA’s large scale and highly imbalanced composition space. {As quantified in Appendix Tab.~\ref{tab:cgqa_freq_stats}, only 20\% of compositions (head) account for 77.3\% of training images, while 50\% of compositions (tail) are observed only once. Moreover, 22.4\% of test images correspond to compositions that never appear in the training set, making CGQA an extremely long-tailed and partially shifted benchmark.} Even with small mini-batches during graph-based prototype updates, the long-tailed frequency of state-object pairs can introduce update bias, slightly skewing prototypes toward frequent compositions and dampening CW improvements.
{Despite this challenge, our frequency-based analysis in Appendix Tab.~\ref{tab:cgqa_freq_perf} shows that Duplex with local graph refinement consistently improves over its ablated variant across head, medium, tail, and unseen groups on CGQA, and also yields higher best HM and AUC, indicating that the method remains beneficial even in the most imbalanced regime.}
By contrast, the OW setting (Tab.~\ref{tab:main_result_open}) plays to the strengths of our approach. Because \emph{Duplex} performs dynamic, semantics-guided prototype refinement rather than relying on fixed compositional priors, it can adapt prototypes on the fly and better disentangle and recombine states and objects when seen and unseen compositions are mixed at test time. This adaptive behavior yields robust generalization.

{
As observed in both the CW setting (Tab.~\ref{tab:main_result_close}) and the OW setting (Tab.~\ref{tab:main_result_open}), accuracy on seen classes (S) consistently exceeds that on unseen classes (U). We attribute this persistent gap to \textbf{visual contextual dependency}. Even when semantic primitives are available during training, their visual manifestations can vary across compositions (e.g., context-dependent state appearances), which hinders generalization to novel visual compositions. Comparing the two tables, the OW setting yields lower absolute scores primarily due to the \textbf{expanded search space}. Unlike the CW setting, which searches only valid pairs, the OW setting considers the full Cartesian product with many invalid distractors, substantially increasing task difficulty.}

{In our experiments, LOGICZSL is the strongest baseline on CGQA, slightly outperforming Duplex on some metrics, whereas Duplex is comparable to or better than LOGICZSL on MIT-States and UT-Zappos under both CW and OW settings. We attribute this pattern to how the two methods exploit label-space structure. LOGICZSL leverages global, logic-induced semantic constraints derived from large language models, which can remain effective even for extremely rare compositions in CGQA’s large and long-tailed label space. Duplex, in contrast, refines a global codebook of image-grounded visual prototypes through label-conditioned mini-batch local graphs and counterfactual compositions. As a result, prototypes for very rare compositions are updated only when those compositions appear during training.  These differences suggest that the approaches are complementary rather than contradictory. LOGICZSL excels when strong global semantic priors are available, while Duplex emphasizes semantically guided refinement of visual prototypes. Integrating logic-induced priors into Duplex in the spirit of LOGICZSL is a promising direction for future work.}

\subsection{Ablation study}

\textbf{Effects of Semantic and Visual Prototype Modules.}
We evaluate the impact of the semantic and visual prototype modules, including three-path prompt learning, the visual prototype (VP) module, and the semantic prototype (SP) module. Tab.~\ref{tab:ablation_LTC} reports results on UT-Zappos and MIT-States.

Row (0) shows that three-path prompts yield the strongest baseline, which we adopt following~\cite{Huang2024Troika}. Rows (1) and (2) indicate that adding either SP or VP individually improves performance. Fig.~\ref{fig:text_vison_embeddingshow} further illustrates their complementarity: visual features (Fig.~\ref{fig:text_vison_embeddingshow}(a),(b)) outperform standalone semantic features in classification, whereas their fusion in \emph{Duplex} (Fig.~\ref{fig:text_vison_embeddingshow}(c)) yields the highest accuracy. Row (3) shows that combining SP and VP further improves unseen accuracy and AUC.
\begin{table}[!htp]
\begin{center}
\caption{Ablate the components in \emph{Duplex} on the CW dataset UT-Zappos and MIT-States. c-s-o denote composition, state, and object prompt, VP denotes the visual prototype module, SP denotes the semantic prototype module.}
\setlength{\tabcolsep}{2pt}
\renewcommand{\arraystretch}{0.8}
\resizebox{0.9\linewidth}{!}{
\begin{tabular}{cccc| 
cccc| 
cccc}
\toprule
\multicolumn{4}{c|}{Module}
& \multicolumn{4}{c|}{ {MIT-States }} 
& \multicolumn{4}{c}{ {UT-Zappos }} \\
& c-s-o & SP & VP
& S & U & HM & AUC 
& S & U & HM & AUC \\
\midrule
 (0) &\Checkmark & \XSolidBrush & \XSolidBrush 
 &48.6 &49.8 &36.8 &20.1
 &66.4 &69.6 &51.1 &37.8 \\ 
 (1)  &\Checkmark &\Checkmark & \XSolidBrush
 & 49.0 & 52.4 & 37.9 & 21.7
 & 66.9 & 74.5 & 54.7 & 42.1 \\
 (2)  &\Checkmark & \XSolidBrush & \Checkmark
 & 49.4 & 52.1 & 38.5 & 21.9
 & 68.1 & 72.9 & 56.3 & 43.7 \\
 (3) &\Checkmark &\Checkmark & \Checkmark
 & {49.7} & {55.6}  & \textbf{40.9} & \textbf{23.7} 
 & {70.5} & {75.6} & \textbf{58.2} & \textbf{46.2} \\
\bottomrule
\end{tabular}
}
\label{tab:ablation_LTC}

\end{center}
\end{table}

\begin{table}[tp]
\begin{center}
\caption{Results on CW datasets (MIT-States and CGQA) under different visual-prototype update strategies. Rows (0) and (1) use fixed momentum values $\lambda=1$ and $\lambda=0$, respectively. Row (2) implements our proposed update, which blends the initialization and batch updates as $\mathcal{N}_{\text{blend}}=\lambda\,\mathcal{N}_{\text{init}}+(1-\lambda)\,\mathcal{N}_{\text{batch}}$, where $\lambda$ is selected on the validation set.}

\setlength{\tabcolsep}{2pt}
\resizebox{0.9 \linewidth}{!}{
\begin{tabular}{cl|
cccc|
cccc}
\toprule
\multicolumn{2}{c|}{\multirow{2}{*}{{Update Strategy}}}
& \multicolumn{4}{c|}{ {MIT-States}} 
& \multicolumn{4}{c}{ {CGQA}} \\
 
& & S & U & HM & AUC 
& S & U & HM & AUC \\
\midrule
 (0) & $ \mathcal{N}_{init} $
 &48.9 &49.9 &37.4 &20.4
 &{41.3} &33.6 &26.6 &11.8 \\ 
 (1)  & $ \mathcal{N}_{batch}$
 &46.8 & 51.1 & 36.9 & 19.9
 & 40.1 & 34.0 & 28.5 & 11.6 \\
 
 (2)  & $\mathcal{N}_{blend}$
 & {49.7} & {55.6} & \textbf{40.9} & \textbf{23.7}
 & 41.1 & {36.2} & \textbf{30.1} & \textbf{13.2} \\
\bottomrule
\end{tabular}
}
\label{tab:ablation_update}
\end{center}
\end{table}

\begin{figure}[pos=t]
    \centering
    \includegraphics[width=1.0\linewidth]{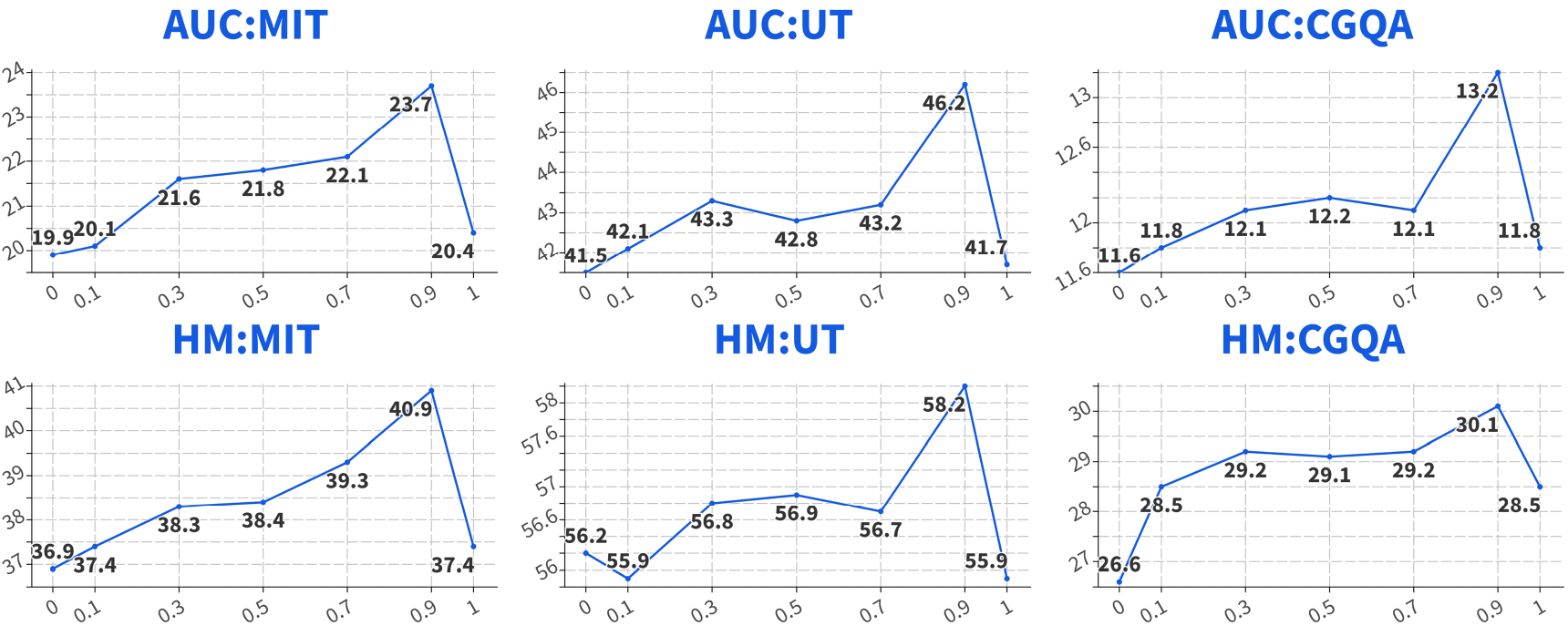}
\caption{Sensitivity analysis of the visual-prototype update coefficient $\lambda$. We report performance on three benchmarks (MIT-States, UT-Zappos, and CGQA) as $\lambda$, which controls the retention ratio of the original visual prototype, is varied. The top row shows AUC, and the bottom row shows HM. {The results consistently indicate that performance peaks in a relatively high-$\lambda$ regime, with $\lambda=0.9$ yielding optimal or near-optimal scores on the three benchmarks shown.}}
    \label{fig:Sensitivity_update}
\end{figure}

\begin{table}[!htb] 
\caption{Results on CW Dataset UT-Zappos. Row (0) corresponds to training with state and object probabilities, while Rows (1)–(2) leverage visual and semantic prototypes, respectively. Row (3) represents our best-performing loss function.}
\small
\renewcommand{\arraystretch}{0.7}
\centering
\resizebox{0.75\linewidth}{!}{
\begin{tabular}{ccc| cccc }
\toprule
\multicolumn{3}{c|}{Loss formulation}
&  S & U & HM & AUC \\
\midrule
 (0) &\multicolumn{2}{l|}{$\mathcal{L}_{s}+\mathcal{L}_{o}$}
 & 30.4 & 54.4 & 30.8 & 13.1\\
 \midrule
 (1) & $\mathcal{L}^{h}_{c}$ &
 &67.2 &68.0 &56.4 &42.0 \\ 
 &   
 \multicolumn{2}{r|}{\textbf{+} $(\mathcal{L}_{s}+\mathcal{L}_{o})$}
  & 68.5 & 73.2  & 57.1 & 44.9 \\
\midrule
 (2) & \multicolumn{2}{l|}{$\mathcal{L}^{t}_{c}$} &
 66.9 & 68.9  & 52.1 & 39.6 \\ 
 
 & \multicolumn{2}{r|}{\textbf{+} $(\mathcal{L}_{s}+\mathcal{L}_{o})$}
 & 68.4 & 71.2 & 54.5 & 40.3\\

  \midrule
  (3) & \multicolumn{2}{l|}{$\mathcal{L}^{h}_{c}+\mathcal{L}^{t}_{c}$}
 & {70.7} & 74.2  & 57.1 & 45.1 \\ 
 
& \multicolumn{2}{r|}{\textbf{+} $(\mathcal{L}_{s}+\mathcal{L}_{o})$}
 & 70.5 & {75.6} & \textbf{58.2} & \textbf{46.2} \\
\bottomrule
\end{tabular}
}
\label{tab:ablation_Loss}
\end{table}

\begin{table}[!htb]    
\centering
\caption{Results on the CW dataset MIT-States under different inference formulations. Rows (0)-(2) use the factorized term $p(s)\,p(o)$, the semantic-prototype term $p(c^{t})$, and the visual-prototype term $p(c^{h})$, respectively. Row (3) corresponds to our full inference formulation and achieves the best performance.}

\small
\renewcommand{\arraystretch}{0.5}
\resizebox{0.8\linewidth}{!}{
\begin{tabular}{ccc| cccc }
\toprule
\multicolumn{3}{c|}{Inference formulation}
&  S & U & HM & AUC \\
\midrule
 (0)  & \multicolumn{2}{l|}{$p(s) \cdot p(o)$}
 & 45.7 & 34.2 & 26.1 & 11.3\\
 \midrule
 (1)  & \multicolumn{2}{l|}{$p(c^{t})$}
 & 50.5 & 47.8 & 36.8 & 20.1\\
  & \multicolumn{2}{r|}{\textbf{+} $p(s) \cdot p(o)$}
 & 49.4 & 51.3 & 37.6 & 21.2\\
 \midrule
 (2) & \multicolumn{2}{l|}{$p(c^{h})$}
 & 42.6 & 45.6 & 32.6 & 15.8\\ 
  & \multicolumn{2}{r|}{\textbf{+ $p(s) \cdot p(o)$}}
 &45.8 &47.1 &33.8 &17.5\\
 \midrule
 (3) & \multicolumn{2}{l|}{$p(c^{h})+p(c^{t})$}
 & {51.2} & 54.0  & 39.3 & 22.9 \\
 
 & \multicolumn{2}{r|}{\textbf{+} $p(s) \cdot p(o)$}
 & 49.7 & {55.6} & \textbf{40.9} & \textbf{23.7} \\
\bottomrule
\end{tabular}
}
\label{tab:ablation_Inference}
\end{table}

\textbf{Effects of Visual Prototype Update.}
We examine the impact of the update rule $\lambda\,\mathcal{N}_{\text{init}} + (1-\lambda)\,\mathcal{N}_{\text{batch}}$ (Tab.~\ref{tab:ablation_update}).
The initialization strategy $\mathcal{N}_{\text{init}}$ computes visual prototypes as the mean of CLIP-encoded training features, whereas $\mathcal{N}_{\text{batch}}$ updates prototypes online by incorporating state and object features extracted from each training mini-batch.
We report the extreme settings $\lambda\in\{0,1\}$. With $\lambda=1$ (pure $\mathcal{N}_{\text{init}}$), we initialize prototypes using averaged CLIP training features before applying the proposed updates. 
With $\lambda=0$ (pure $\mathcal{N}_{\text{batch}}$), prototypes are randomly initialized and updated solely using state and object features present in each mini-batch.
Across datasets, the optimal fixed $\lambda$ varies. On MIT-States and CGQA, using either $\mathcal{N}_{\text{init}}$ or $\mathcal{N}_{\text{batch}}$ alone underperforms the combined approach, except for a slightly higher seen metric on CGQA with $\mathcal{N}_{\text{init}}$. Consequently, selecting $\lambda$ based on validation performance yields the best results; for MIT-States, $\lambda=0.9$ is optimal.

\begin{table*}[htb]
\renewcommand{\arraystretch}{0.9}
\centering
\caption{Quantitative results in the \textbf{CW} setting on three benchmarks using \textbf{ResNet-18} and \textbf{BLIP} backbones. \emph{CGE}~\cite{naeem2021learning} and \emph{Semantic Proto} are the base models. ``\textbf{$+$ Visual Proto}'' adds our dynamic visual-prototype refinement, and ``\textbf{$+$ Visual Proto (\emph{Duplex})}'' denotes the full model with semantic-visual synergy. Best results are in \textbf{bold}.}

\begin{center}
\resizebox{0.9\linewidth}{!}{
\begin{tabular}{c|l|
ccc c|
ccc c|
ccc c}
\toprule
\multirow{2}{*}{Backbone} & \multicolumn{1}{c|}{\multirow{2}{*}{Method}}
& \multicolumn{4}{c|}{{MIT-States}} 
& \multicolumn{4}{c|}{{UT-Zappos}} 
& \multicolumn{4}{c}{{CGQA}} \\
&
& S & U & HM & AUC 
& S & U & HM & AUC 
& S & U & HM & AUC \\

\midrule

& CGE~\cite{naeem2021learning}
& 32.8 & 28.0 & 21.4 & 6.5 
& {64.5} & 71.5 & 60.5 & 33.5 
& 31.4 & 14.0 & 14.5 & 3.6 \\

\multirow{-2}{*}{\cellcolor{white}ResNet-18} & \textbf{+ Visual Proto}
&{32.8} & {29.3} & \textbf{22.6} & \textbf{7.4} 
& 63.2 & {72.3} & \textbf{61.2} & \textbf{34.4} 
& {32.0} & {21.8} & \textbf{18.7} & \textbf{5.4} \\

\midrule
& Semantic Proto
& 54.7 & 49.2 & 37.2 & 20.7 
& 63.4 & 68.3 & 52.3 & 37.8 
& {43.1} & 33.3 & 28.2 & 11.8 \\

\multirow{-2}{*}{\cellcolor{white}BLIP} & \textbf{+ Visual Proto (\emph{Duplex})}
& {57.7} & {51.7} & \textbf{41.0} & \textbf{24.6} 
& {70.9} & {75.0} & \textbf{57.9} & \textbf{46.1} 
& 41.7 & {37.5} & \textbf{31.4} & \textbf{14.0} \\

\bottomrule
\end{tabular}}
\label{tab:main_result_close_combined}
\end{center}
\end{table*}

\begin{table}[th]
\centering
\setlength{\tabcolsep}{2pt}
\renewcommand{\arraystretch}{1.0}

\caption{
Efficiency and accuracy comparison on UT-Zappos~\cite{UT-Zappos}. 
C-S-O denotes the CLIP-based baseline. 
We report trainable parameters, peak GPU memory usage, and runtime (training time per epoch and total inference time on the test set), together with CW AUC/HM metrics. L.G.\ and G.C.\ denote Local Graph and Global Codebook. All models use the same batch size ($B=64$) and optimizer settings.
}

\resizebox{1.0\linewidth}{!}{
\begin{tabular}{l|cc|ccc}
\toprule
{Method} & {L.G.} & {G.C.} & {Params/Peak Mem$\downarrow$} & {Train/Test time}$\downarrow$ & {AUC/HM$\uparrow$} \\
\midrule
Troika~\cite{Huang2024Troika}& \XSolidBrush & \XSolidBrush & 22.0M/20.3G & 12.4min/51s & 41.9/54.6 \\
C-S-O & \XSolidBrush & \XSolidBrush & 8.9M/17.8G & 12.0min/46s & 37.8/51.1 \\
\emph{Duplex}& \XSolidBrush & \Checkmark & 9.4M/21.1G & 16.4min/46s & 42.6/56.0 \\
\textbf{\emph{Duplex}}& \Checkmark & \Checkmark & 9.5M/21.1G & 16.6min/47s & \textbf{46.2/58.2} \\
\bottomrule
\end{tabular}}
\label{tab:efficiency}

\end{table}

\begin{table}[thb]
\renewcommand{\arraystretch}{0.5}

    \centering
    \caption{\textbf{Effects of semantic anchors on visual refinement on UT-Zappos.} The visual graph provides limited gains with fixed templates, but yields substantial improvements when paired with soft prompts.}
    \label{tab:ablation_components}
    \resizebox{1.0\linewidth}{!}{
\begin{tabular}{cc | c c c c }
        \toprule
        \multicolumn{2}{c|}{\textbf{Components}} & \multicolumn{4}{c}{\textbf{UT-Zappos}} \\
        Soft Prompt & Visual Prototype & S & U & HM & AUC \\
        \midrule
        \XSolidBrush & \XSolidBrush & 63.5 & 68.2 & 49.6 & 38.0 \\
        \XSolidBrush & \Checkmark & 63.8 & 68.5 & 49.9 & 38.4 \\
        \midrule
        \Checkmark & \XSolidBrush & 66.9 & 74.5 & 54.7 & 42.1 
        \\
        \Checkmark & \Checkmark & \textbf{70.5} & \textbf{75.6} & \textbf{58.2} & \textbf{46.2} \\
        \bottomrule
    \end{tabular}
    }

\end{table}

\textbf{Sensitivity of Hyperparameter $\lambda$.}
In Fig.~\ref{fig:Sensitivity_update}, we examine how varying the update coefficient $\lambda$, which regulates the update rate for each composition, affects performance. Across all three datasets, performance improves as $\lambda$ increases and peaks around $\lambda=0.9$ in terms of AUC. When $\lambda$ becomes larger, performance declines as updates rely increasingly on CLIP-initialized features, which slows the adaptation of state and object node features and reinforces pretraining alignment, limiting refinement. 
{Based on this sensitivity analysis, we tune $\lambda$ on a validation split for each dataset within the range $[0, 1]$. The selected values typically fall in a relatively high range, with $\lambda=0.9$ being optimal or near-optimal on the benchmarks shown in Fig.~\ref{fig:Sensitivity_update}, which motivates the choice used in our main results.}

\textbf{Effects of Inference Formulation.}
We examine the impact of the inference formulation $p(c^{h}) + p(c^{t}) + p(s)\,p(o)$, as reported in Tab.~\ref{tab:ablation_Inference}. Here, $p(c^{h})$ and $p(c^{t})$ denote composition-level predictions from the \emph{visual} and \emph{semantic} prototype branches, respectively, while $p(s)\,p(o)$ represents the factorized joint probability of state and object predictions. Either $p(c^{h})$ or $p(c^{t})$ alone yields strong performance; however, results on MIT-States indicate that $p(s)\,p(o)$ alone is an unreliable predictor. Crucially, adding $p(s)\,p(o)$ to the composition-only combination $p(c^{h}) + p(c^{t})$ further improves unseen accuracy. These findings highlight that combining semantic and visual composition prototypes, augmented by factorized state--object evidence, improves generalization in zero-shot compositional learning.

\textbf{Effects of Loss Formulation.}
Tab.~\ref{tab:ablation_Loss} reports results for different loss formulations on UT-Zappos under the closed-world (CW) setting.
Row (0) is a baseline trained with only state and object classification losses, whereas Rows (1)-(2) progressively add the visual-prototype and semantic-prototype losses, respectively. 
Row (3) (our final formulation) jointly optimizes composition losses at both the visual and semantic levels, together with state and object regularization, achieving the best overall performance (\textbf{HM} = 58.2, \textbf{AUC} = 46.2).
These results indicate that the proposed composite loss enhances the discriminability of both visual and semantic prototypes and better captures the joint influence of state-object pairs within each image, thereby improving compositional recognition accuracy.

\textbf{Experiments with Other Backbones.}
To demonstrate the generality of our approach, we conduct additional experiments with backbones beyond CLIP. Specifically, we consider \textbf{ResNet-18} and \textbf{BLIP} as representative architectures. For the ResNet-18 backbone, we implement our method on top of CGE~\cite{naeem2021learning}, which uses ResNet-18 and, crucially, focuses solely on learning semantic composition representations. This makes CGE an ideal baseline for isolating the contribution of our visual-prototype branch. We also instantiate our method by replacing CLIP's text and image encoders with their BLIP counterparts. The corresponding results are reported in Tab.~\ref{tab:main_result_close_combined}.

\textbf{Efficiency Analysis.}
Tab.~\ref{tab:efficiency} compares efficiency and accuracy on UT-Zappos for Troika~\cite{Huang2024Troika}, the C-S-O baseline, and our Duplex variants. All methods are based on CLIP ViT-L/14 with a batch size of 64 and identical optimizer settings.
We explicitly disentangle the contributions of the \emph{Local Graph} (L.G.) and the \emph{Global Codebook} (G.C.), and note that the local graph is instantiated only during training for prototype refinement and is not used at test time.

In terms of model capacity and deployment cost, Duplex remains lightweight. Compared to C-S-O (8.9M parameters), enabling the global visual codebook in Duplex (without the local graph) increases the parameter count only slightly to 9.4M, and full Duplex with L.G.\ uses 9.5M parameters. Both are far below Troika's 22.0M. Importantly, inference-time cost is almost unchanged with and without the local graph. Both C-S-O and Duplex without L.G.\ require 46~s to process the test set, and full Duplex requires 47~s, which is still faster than Troika (51~s). This aligns with our design, since Duplex performs no graph construction during inference and uses fixed prototypes for scoring.

In terms of memory and training overhead, Duplex incurs a moderate but controlled cost. Peak GPU memory increases from 17.8~GB (C-S-O) to 21.1~GB after enabling the global codebook, and remains at 21.1~GB after enabling the local graph. This indicates that the additional memory is primarily due to storing the global codebook and CLIP activations rather than graph construction. Training time per epoch increases from 12.0~min (C-S-O) to 16.4~min for Duplex without L.G.\ and to 16.6~min for full Duplex. This reflects the cost of prototype refinement while remaining comparable to Troika's 12.4~min.

Despite this moderate training-time overhead, Duplex yields clear accuracy gains. Adding the global codebook alone improves CW AUC/HM from 37.8/51.1 (C-S-O) to 42.6/56.0, slightly surpassing Troika (41.9/54.6). Enabling the local graph further improves performance to 46.2/58.2, achieving the highest accuracy among all compared methods. Overall, the additional components (G.C.\ and L.G.) provide commensurate benefits in compositional recognition, while keeping the parameter count and inference-time complexity comparable to the CLIP baseline.

\textbf{Effects of Semantic Anchors on Visual Refinement.} 
To explicitly validate the synergy between soft prompt learning and the dynamic visual graph, we conduct a component-wise ablation study on UT-Zappos, as reported in Tab.~\ref{tab:ablation_components}. We examine whether the effectiveness of the dynamic visual graph depends on the stability of semantic anchors. As shown in the first two rows, when the model uses \textit{Fixed Templates}, introducing the visual graph yields only a negligible AUC improvement (from 38.0\% to 38.4\%). This suggests that fixed templates suffer from domain misalignment, producing inaccurate anchors that fail to guide graph-based feature propagation effectively.

In contrast, when paired with \textit{Soft Prompts} (last two rows), the visual graph delivers substantial gains, increasing AUC from 42.5\% to 46.2\% ($+3.7\%$). This interaction confirms that soft prompts play a dual role in \emph{Duplex}. They not only enhance semantic expressiveness but also provide stable semantic anchors that better condition the visual graph updates. Such stability is necessary for the dynamic visual graph to regularize visual features without inducing semantic drift.

\begin{figure*}[thb]
\centering
\includegraphics[width=\linewidth]{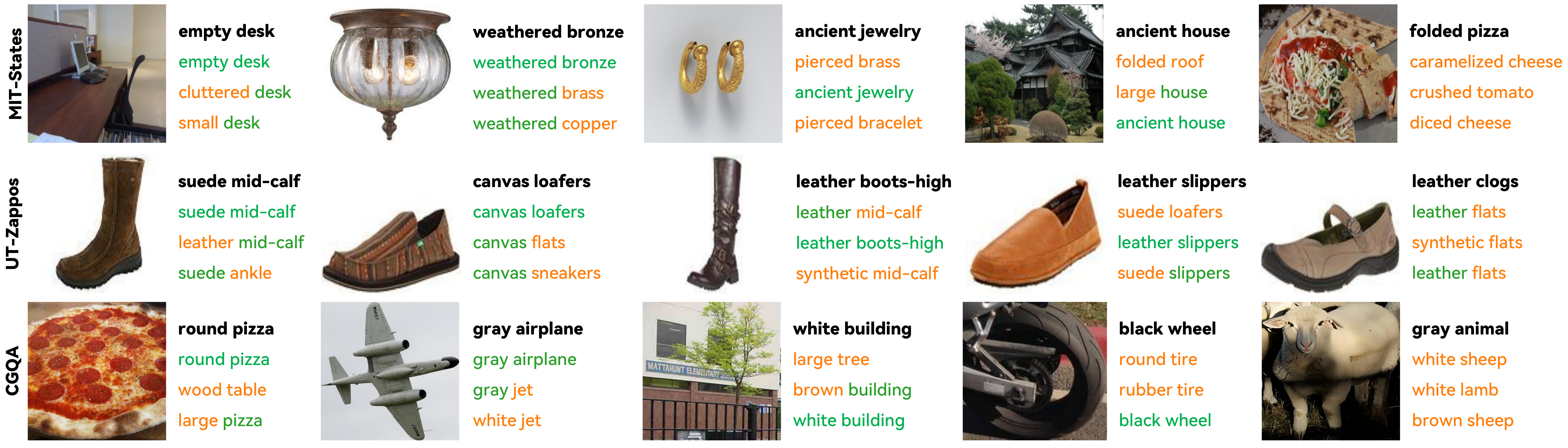}
\caption{\textbf{Qualitative Results}. We randomly sample examples from MIT-States (top row), UT-Zappos (middle row), and CGQA (bottom row). Each image shows the ground-truth label (black) and the \textit{top-3} predictions (colored), with correct predictions highlighted in green.}
\label{fig:success_case}
\end{figure*}

\begin{figure*}[!hb]
\centering
\includegraphics[width=0.9\linewidth]{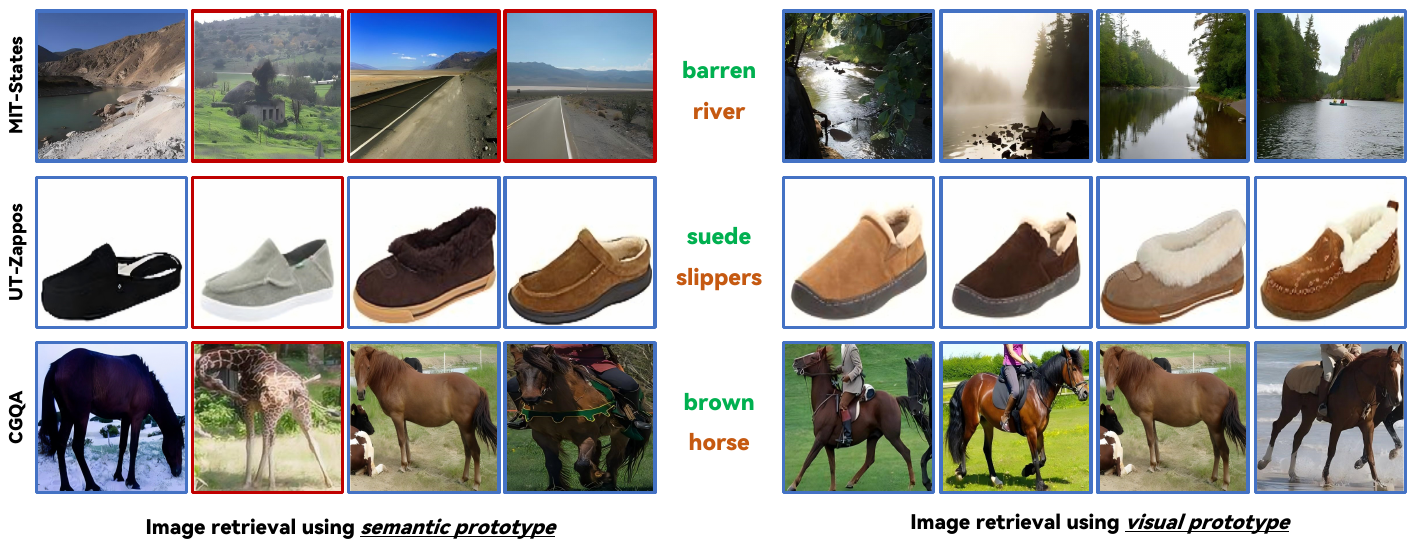}
\caption{\textbf{Semantic and Visual Prototype Retrieval.} We perform retrieval using semantic and visual prototypes extracted by \emph{Duplex} across three datasets. Incorrect results are highlighted in red: \textit{barren road} in the first row, \textit{canvas loafers} in the second, and \textit{brown giraffe} in the third.}

\label{fig:text_vision_Retrieval}
\end{figure*}

\subsection{Qualitative Results}
Inspired by prior work~\cite{hao2023ade,2022ECCV_Learning_Invariant_Visual,Saini_2022_CVPR}, we conduct a qualitative analysis of image--text retrieval to illustrate how \emph{Duplex} aligns images and text.

\textbf{Image-to-Composition Retrieval.}
Fig.~\ref{fig:success_case} presents qualitative results on the test sets of MIT-States, UT-Zappos, and CGQA, covering both seen and unseen compositions.  Given an input image (e.g., ``\textit{ancient house}''), we extract its visual-prototype feature and retrieve the top three most similar \emph{textual} composition embeddings. Although the exact label does not always appear among the top-ranked predictions, the retrieved compositions are often semantically relevant. For example, in the ``\textit{folded pizza}'' image (row~1, column~4), the exact label is absent from the top three matches, yet the retrieved compositions (``\textit{crushed tomato}'' and ``\textit{diced cheese}'') are reasonable and visually supported. A similar pattern is observed for the ``\textit{gray animal}'' image (row~3, column~4), where the retrieved compositions differ from the label but remain semantically meaningful. These observations suggest that our model captures coherent visual-semantic relations beyond exact-match accuracy.

\textbf{Semantic and Visual Retrieval.}
We evaluate semantic-to-visual and visual-to-image retrieval with \emph{Duplex}. In semantic retrieval, a semantic prototype (e.g., ``\textit{Suede Slippers}'', row~2) is used to retrieve the four most similar images (left panel of Fig.~\ref{fig:text_vision_Retrieval}). In visual retrieval (right panel), we retrieve images based on visual-prototype similarity. Overall, visual retrieval attains higher precision, whereas semantic retrieval is more prone to mismatches (e.g., ``\textit{Suede Slippers}'' retrieving ``\textit{Canvas Loafers}''). A similar pattern holds for ``\textit{Brown Horse}'', underscoring the effectiveness of visual prototypes in capturing compositional structure while mitigating semantic bias.

\begin{figure}[ht]
\centering
  \includegraphics[width=\linewidth]{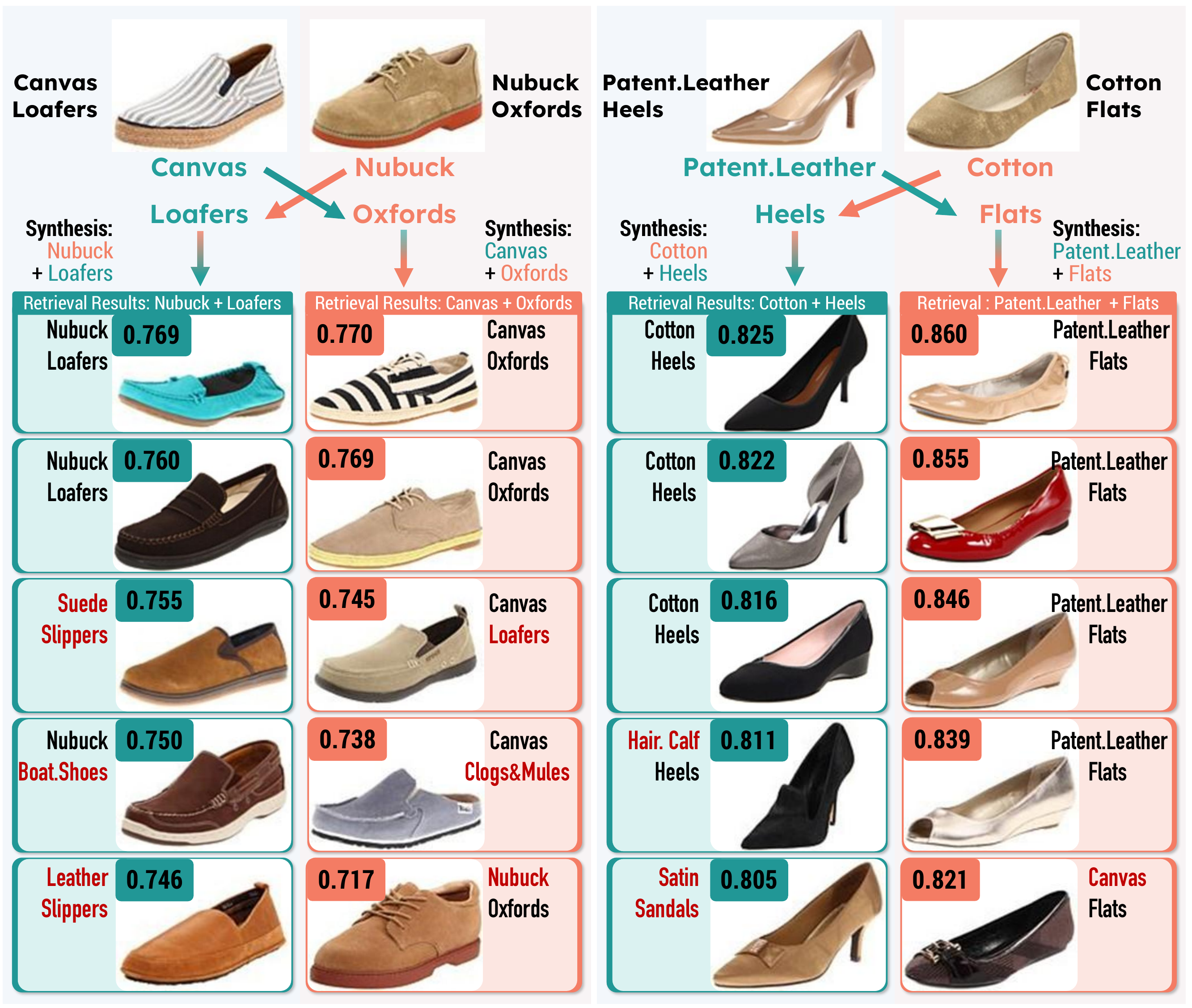} 
\caption{{\textbf{Feature Swapping Retrieval on UT-Zappos.} We evaluate disentanglement by swapping state and object features between two source images. The ``Synthesis'' arrow indicates the combination of the state feature from one source and the object feature from the other. The table reports the top-5 retrieved images based on the synthesized features. The model achieves accurate \textbf{Top-1 retrieval} (e.g., 0.769 for ``Nubuck Loafers''), demonstrating effective functional disentanglement, while minor state drift in lower ranks (e.g., top-3 to top-5) highlights the challenge of object dominance in visual semantics.}}
\label{fig:swapping}
\vspace{-5mm}
\end{figure}

\textbf{Evaluation of Disentanglement via Feature Swapping.}
To further assess the quality of the learned representations, we conduct a feature-swapping retrieval experiment on the UT-Zappos dataset, following the visualization protocol in prior work~\cite{2025zhangPAMI}. Given two real images, we extract their state and object representations, swap them to synthesize novel feature compositions, and retrieve the top-5 nearest neighbors from the test set based on feature distance. This task is more challenging than standard state-object retrieval, as it requires precise separation and recombination of visual semantics without direct ground-truth supervision for the synthesized pairs. Fig.~\ref{fig:swapping} presents qualitative results. The top row shows the source image pairs, and the lists below show the top-5 neighbors retrieved using the synthesized features.

\noindent\textbf{Success on Top-1 Retrieval.}
In the left panel, we synthesize a query by combining the visual state ``Nubuck'' (extracted from \textit{Nubuck Oxfords}) and the visual object ``Loafers'' (extracted from \textit{Canvas Loafers}). The model successfully retrieves ``Nubuck Loafers'' as the exact Top-1 match, with a cosine similarity of \textbf{0.769}. Similarly, the right panel shows correct retrieval of ``Patent Leather Flats'' (Top-1 similarity: \textbf{0.860}) by combining features from \textit{Heels} and \textit{Flats}. Notably, Top-1 retrieval does not collapse to either the source object (e.g., \textit{Oxfords}) or the source state (e.g., \textit{Canvas}), indicating that the learned factors are compositional and suppress source-specific identity.

\noindent\textbf{Analysis of Object Dominance.}
While disentanglement remains robust for the most relevant match (Top-1), we observe ``object dominance'' in lower-ranked results (e.g., ranks 3-5). In these cases, the retrieved images consistently match the target object but may drift in the state. This suggests that object features, largely determined by rigid shape, provide a stronger and more stable cue, whereas state features, often reflected in texture or material, are visually subtler and harder to preserve as confidence decreases. Reducing this imbalance and further enhancing state distinctiveness is an important direction for future work.

\begin{figure*}[htb] 
\centering 
\includegraphics[width=0.9\textwidth]{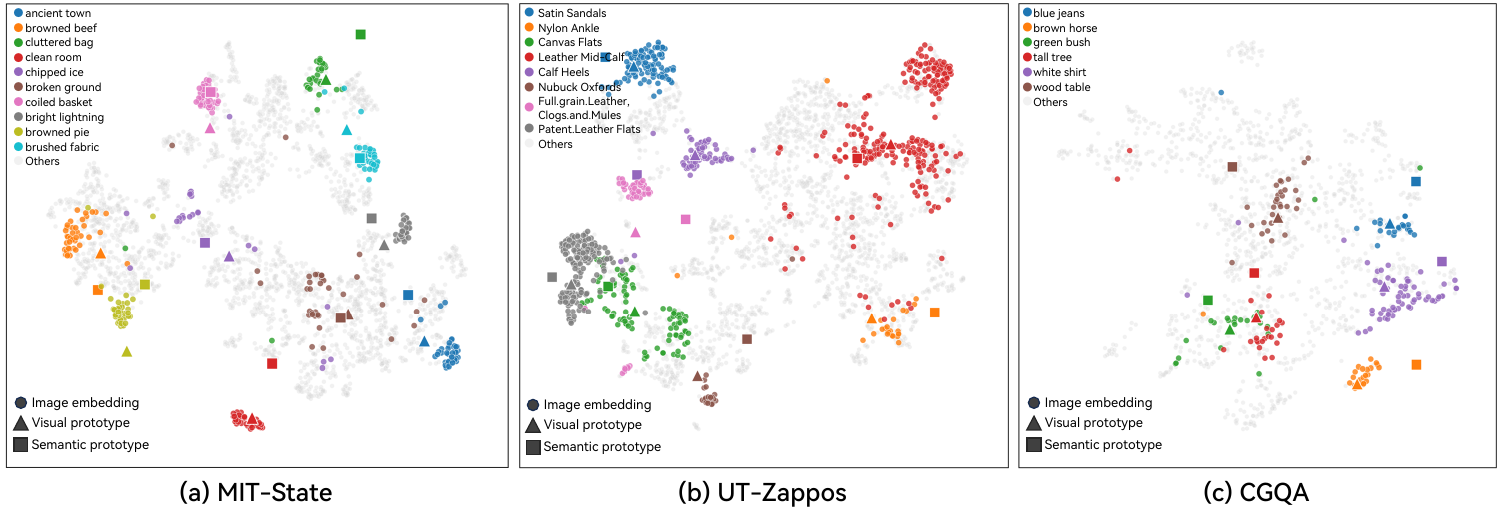}
\caption{t-SNE visualization of visual prototypes ($\triangle$), semantic prototypes ($\Box$), and test-sample embeddings ($\circ$) on (a) MIT-States, (b) UT-Zappos, and (c) CGQA, including both seen and unseen compositions.}
\label{fig:visual_text_prototype}
\end{figure*}

\begin{figure}[pos=htb] 
\centering 
\includegraphics[width=\linewidth]{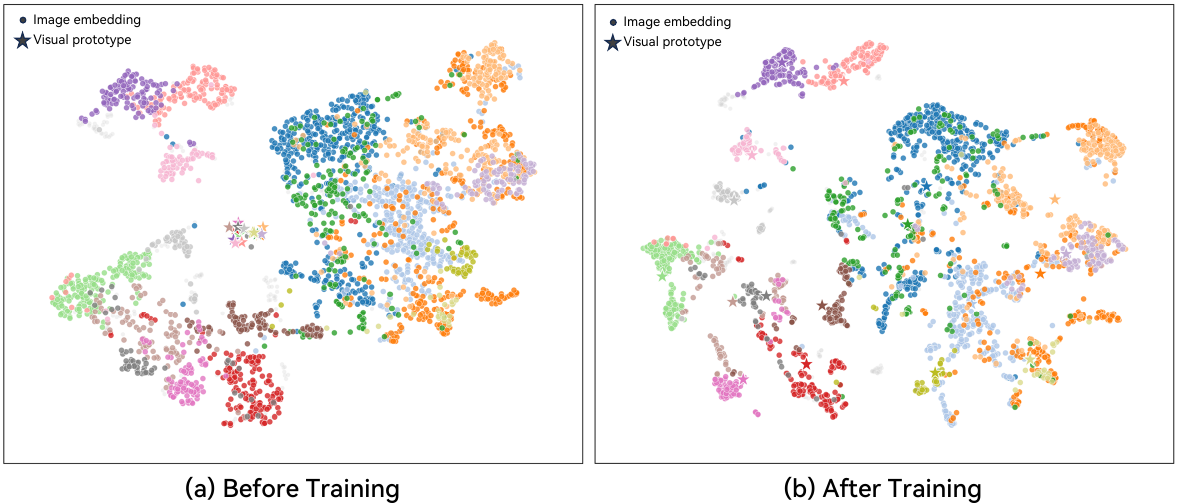}
\caption{t-SNE visualization of UT-Zappos test samples before and after training (zoom in for details).}
\label{fig:training_before_after}
\end{figure}

\textbf{Visualization in Latent Space.}
To facilitate qualitative analysis, Fig.~\ref{fig:visual_text_prototype} visualizes visual and semantic prototypes across three datasets, together with test-sample embeddings. Colors denote composition categories. The visual prototypes ($\triangle$) tend to lie closer to the centers of test-sample clusters ($\circ$), whereas the semantic prototypes ($\Box$) are generally farther away. Nevertheless, semantic prototypes provide complementary signals. For example, on MIT-States, the category shown in pink has its semantic prototype closer to the cluster center. These visualizations suggest that \emph{Duplex} learns robust composition representations.

Fig.~\ref{fig:training_before_after} shows 2D embeddings of UT-Zappos test samples produced by \emph{Duplex}. Each point represents an image, and {\small\FiveStar} marks the visual prototype. Colors correspond to categories, and we visualize the 20 most populous categories. In composition classification, better-separated embeddings are typically more discriminative. The proximity of {\small\FiveStar} to a cluster center reflects how well the visual prototype anchors its composition. Compared with pre-training, the trained \emph{Duplex} model exhibits tighter clusters and improved alignment between prototypes and sample embeddings. This improvement stems from \emph{Duplex}'s dynamic updates of visual prototypes, which yield more discriminative composition features. Overall, visual prototypes enhance the model's ability to discriminate between categories. However, \emph{Duplex} still struggles with tightly coupled categories, highlighting an important direction for future improvement.

\textbf{Limitations.}
We observe substantial performance improvements on smaller-scale or domain-specific datasets (e.g., UT-Zappos), likely due to reduced uncertainty within these specialized domains and more balanced composition distributions.
However, on large-scale, general-purpose datasets (e.g., CGQA), the gains are relatively modest.
{As analyzed in Appendix Tab.~\ref{tab:cgqa_freq_stats}, CGQA exhibits an extremely long-tailed composition distribution, where a small head of frequent compositions dominates the training instances while most compositions are observed only a few times or remain completely unseen.}
{Our frequency-based analysis in Appendix Tab.~\ref{tab:cgqa_freq_perf} further shows that Duplex consistently improves over its ablated variant without the local graph across head, medium, tail, and unseen compositions, but absolute performance on the rarest compositions remains low due to limited visual supervision.}
Notably, in many misclassified cases, the correct label is ranked among the top three predictions or can be expressed by a semantically equivalent description (Fig.~\ref{fig:success_case}).
We attribute these errors in part to ambiguously labeled samples, which hinder effective refinement of visual prototypes and degrade performance, especially for long-tailed and visually ambiguous compositions.
Future work will explore data and model uncertainty to further enhance compositional learning and develop long-tail-aware refinement strategies (e.g., frequency-aware sampling) tailored to highly imbalanced benchmarks.

\section{Conclusion}
\label{conc}

In this paper, we presented \emph{Duplex}, a compositional zero-shot learning framework that addresses two persistent bottlenecks: semantic projection bias and seen-dominant optimization. Our approach couples dual-prototype learning with active local-graph refinement of visual prototypes. \emph{Duplex} preserves prompt-learned semantic prototypes as stable, interpretable anchors and seeds the visual prototypes by disentangling and counterfactually recombining seen states and objects. A label-conditioned local graph, constructed per mini-batch, enables lightweight message passing under semantic-consistency constraints; this injects fine-grained visual evidence into the prototype space and brings unseen compositions into the training dynamics. At inference, \emph{Duplex} retrieves from a global codebook without additional graph construction. Empirically, \emph{Duplex} achieves competitive performance on MIT-States, UT-Zappos, and CGQA under closed- and open-world settings. Analyses show that semantically guided local updates reduce intra-class variance without collapsing inter-class margins, suppress cross-class shortcuts, and attenuate seen bias, thereby narrowing the train-test gap that often limits VLM-based CZSL. These results support our central claim: refining only the visual prototypes, while using semantic prototypes as anchors, yields more discriminative and adaptable compositional representations.

Although effective, \emph{Duplex} depends on the quality of factor disentanglement and the fidelity of counterfactual recomposition; noise can propagate through the local graph. The method may also be sensitive to prompt design and mini-batch graph sparsity when scaling to very large composition vocabularies. In future work, we will design stronger causal and disentanglement priors and develop uncertainty-aware updates to further stabilize refinement.

\appendix
\section*{Appendix}

\textbf{Composition Frequency Statistics on CGQA.}
We begin by reporting the distribution of state--object compositions across the training and test sets in CGQA. The results are summarized in Tab.~\ref{tab:cgqa_freq_stats}, where we group compositions into three frequency categories: head (most frequent), medium (moderately frequent), and tail (least frequent). We also report unseen compositions that do not appear in the training set but are present in the test set. Specifically, the head category (top 20\% most frequent compositions) accounts for 77.3\% of the training instances, indicating strong concentration in a small subset of compositions. The medium category (next 30\%) comprises 12.3\% of the training instances. The tail category (remaining 50\%) accounts for 10.4\% of the training set, with many compositions observed only once. Finally, unseen compositions represent 22.4\% of the test instances, introducing additional challenges for generalization.

\begin{table}[ht]
\centering

\caption{Composition-level frequency statistics on CGQA, using train-defined head/medium/tail splits. We report the number and percentage of instances in each group for the training and test sets.}
\setlength{\tabcolsep}{6pt}
\resizebox{0.9\linewidth}{!}{
\begin{tabular}{lrrrr}
\toprule
\multirow{2}{*}{Category} & \multicolumn{2}{c}{Train} & \multicolumn{2}{c}{Test} \\
\cmidrule(lr){2-3}\cmidrule(lr){4-5}
 & Count & Instance \% & Count & Instance \% \\
\midrule
Head   & 20{,}821 & 77.3\% & 3{,}309 & 64.9\% \\
Medium &  3{,}303 & 12.3\% &   411   &  8.1\% \\
Tail   &  2{,}796 & 10.4\% &   234   &  4.6\% \\
Unseen &      0   &  0.0\% & 1{,}144 & 22.4\% \\
\midrule
Total  & 26{,}920 & 100.0\% & 5{,}098 & 100.0\% \\
\bottomrule
\end{tabular}
}
\label{tab:cgqa_freq_stats}    

\end{table}

\textbf{Frequency-Based Performance Analysis on CGQA.}
We evaluate \emph{Duplex} with and without the local graph (L.G.) on each frequency group and report instance-level accuracy, together with the corresponding gains brought by the local graph. As shown in Tab.~\ref{tab:cgqa_freq_perf}, \emph{Duplex} with the local graph consistently outperforms its variant without the local graph across all frequency groups. Notably, the head group shows a 1.00\% improvement in instance-level accuracy, while the medium and tail groups improve by 1.21\% and 0.43\%, respectively. Unseen compositions also benefit, with a 1.50\% gain in accuracy. These results show that the local graph improves performance across all frequency groups, including rare and unseen compositions, supporting its robustness.

\begin{table}[ht]
\centering

\caption{Frequency-based performance analysis on CGQA. Instance-level accuracy (\%) of Duplex with and without the mini-batch local graph (L.G.), grouped by training-frequency bins and unseen compositions.}
\setlength{\tabcolsep}{6pt}
\resizebox{0.9\linewidth}{!}{
\begin{tabular}{lrrrr}
\toprule
\multirow{2}{*}{Group} & \multirow{2}{*}{\#Samples} & \multicolumn{2}{c}{Accuracy (\%)} & \multirow{2}{*}{$\Delta$} \\
\cmidrule(lr){3-4}
 &  & w/o L.G. & Duplex &  \\
\midrule
 Head   & 3{,}309 & 45.60 & 46.60 & +1.00 \\
 Medium &   411   & 13.87 & 15.08 & +1.21 \\
 Tail   &   234   &  8.97 &  9.40 & +0.43 \\
\midrule
Seen	& 3,954	& 40.13	& 41.11	& +0.98
\\
Unseen & 1{,}144 & 34.70 & 36.20 & +1.50 \\

\bottomrule
\end{tabular}
}
\label{tab:cgqa_freq_perf}

\end{table}

\printcredits

\bibliographystyle{elsarticle-num-names}

\bibliography{reference}

\end{document}